\documentclass{article}

\PassOptionsToPackage{numbers, compress}{natbib}
\usepackage[main, final]{neurips_2026}

\usepackage[utf8]{inputenc}
\usepackage[T1]{fontenc}
\usepackage[colorlinks=true, citecolor=blue, linkcolor=blue, urlcolor=blue]{hyperref}
\usepackage{url}
\usepackage{booktabs}
\usepackage{amsfonts}
\usepackage{amsmath}
\usepackage{amssymb}
\usepackage{amsthm}
\usepackage{microtype}
\usepackage{xcolor}
\usepackage{graphicx}
\usepackage{algorithm}
\usepackage{algpseudocode}
\usepackage{multirow}
\usepackage{xspace}
\usepackage{enumitem}

\usepackage{caption}
\newcommand{\ours}{\textsc{GazeFlow}}
\newcommand{\R}{\mathbb{R}}

\ifx\figurename\undefined \def\figurename{Figure}\fi
\renewcommand{\figurename}{Figure.}

\newcommand{\Sect}[1]{Section~\ref{#1}}
\newcommand{\Fig}[1]{Figure~\ref{#1}}
\newcommand{\Tbl}[1]{Table~\ref{#1}}
\newcommand{\Equ}[1]{Eq.~\ref{#1}}
\newcommand{\Eqn}[1]{Eq.~\ref{#1}}
\newcommand{\Apx}[1]{Appendix~\ref{#1}}

\newcommand{\proj}{\textsc{GazeFlow}\xspace}

\makeatletter
\renewcommand{\paragraph}{%
  \@startsection{paragraph}{4}%
  {\z@}{0.5ex \@plus 0.5ex \@minus .2ex}{-0.5em}%
  {\normalfont\normalsize\bfseries}%
}
\makeatother

\title{\proj: From Human Gaze Behavior to Generative Egocentric Gaze Prediction}

\author{%
  Sheng Zhao \\
  Department of Computer Science \\
  University of Rochester \\
  \texttt{szhao32@ur.rochester.edu}
  \And
  Weikai Lin \\
  Department of Computer Science \\
  University of Rochester \\
  \texttt{wlin33@ur.rochester.edu}
  \And
  Yuhao Zhu \\
  Department of Computer Science \\
  University of Rochester \\
  \texttt{yzhu@rochester.edu}
}

\begin{document}
\maketitle
\begin{abstract}
Egocentric gaze prediction enables many downstream applications but remains challenging, as human gaze is inherently stochastic.
This stochasticity is constrained by structured temporal dynamics alternating between fixations and saccades, top-down influences from tasks, and bottom-up visual saliency.
Based on this observation, we introduce \proj{}, a framework that directly models gaze as a joint distribution of temporal gaze positions conditioned upon both top-down and bottom-up information. 
In particular, \proj{} uses conditional flow matching (CFM): a learned velocity field iteratively transports a Gaussian noise sample into a plausible gaze trajectory drawn from this joint distribution.
The velocity field is conditioned on bottom-up visual features extracted by a video encoder and on top-down task information obtained by globally querying these features.
On standard datasets, \proj achieves state-of-the-art performance on per-frame metrics, and the generated trajectories align better with human gaze temporal dynamics.
The project page and source code are at: \href{https://horizon-lab.org/gazeflow/}{https://horizon-lab.org/gazeflow/}.
\end{abstract}

\section{Introduction}
\label{sec:intro}

The eye provides a direct window into the brain. 
Egocentric video captures visual experience from the wearer’s perspective, offering a natural way to probe this link---provided that gaze is known.
However, gaze is not directly observable in such videos without dedicated eye-tracking hardware, which requires calibration and introduces additional cost and complexity in real-world deployment.
At the same time, access to gaze would enable a range of applications, including foveated video compression~\citep{itti2004automatic, kaplanyan2019deepfovea, mazumdar2019perceptual, geisler1998real, leng2019energy}, hazard anticipation in autonomous driving~\citep{palazzi2018predicting}, power-efficient rendering~\citep{lin2025powergs, lin2026lowpowar}, and imitation learning in robotics~\citep{kim2021gaze}. 
A natural alternative is to infer where the wearer looks directly from the video itself, a problem known as egocentric gaze prediction~\citep{huang2018predicting,lai2022glc,li2025egom2p}.

Gaze patterns are notoriously difficult to predict because of the stochasticity, as widely recognized in prior work~\cite{lai2022glc, huang2018predicting, li2025egom2p, zhao2025coordauth}: even for the same observer viewing the same scene, trajectories are not exactly reproducible.
This stochasticity, however, is not purely random; rather, it is jointly constrained by three factors.
First, human gaze trajectories exhibit structured dynamics under the oculomotor control, characterized by alternating cycles of saccades and fixational eye movements superimposed with
occasional microsaccades~\cite{rucci2018temporal};
this imposes strong temporal constraints on how gaze can evolve over time.
Second, gaze is strongly task-dependent~\cite{tatler2011eye,hayhoe2005eye, rothkopf2007task, henderson2017gaze};
information about the task, often revealed in future frames, can therefore inform where the observer is likely to look at the current moment.
Third, gaze is also influenced by stimulus-driven factors such as conspicuity and visual saliency~\cite{itti2000saliency, treisman1980feature}, which requires analyzing the spatio-temporal signal properties in the scene.

\begin{figure}[t]
  \centering
  \includegraphics[width=\linewidth]{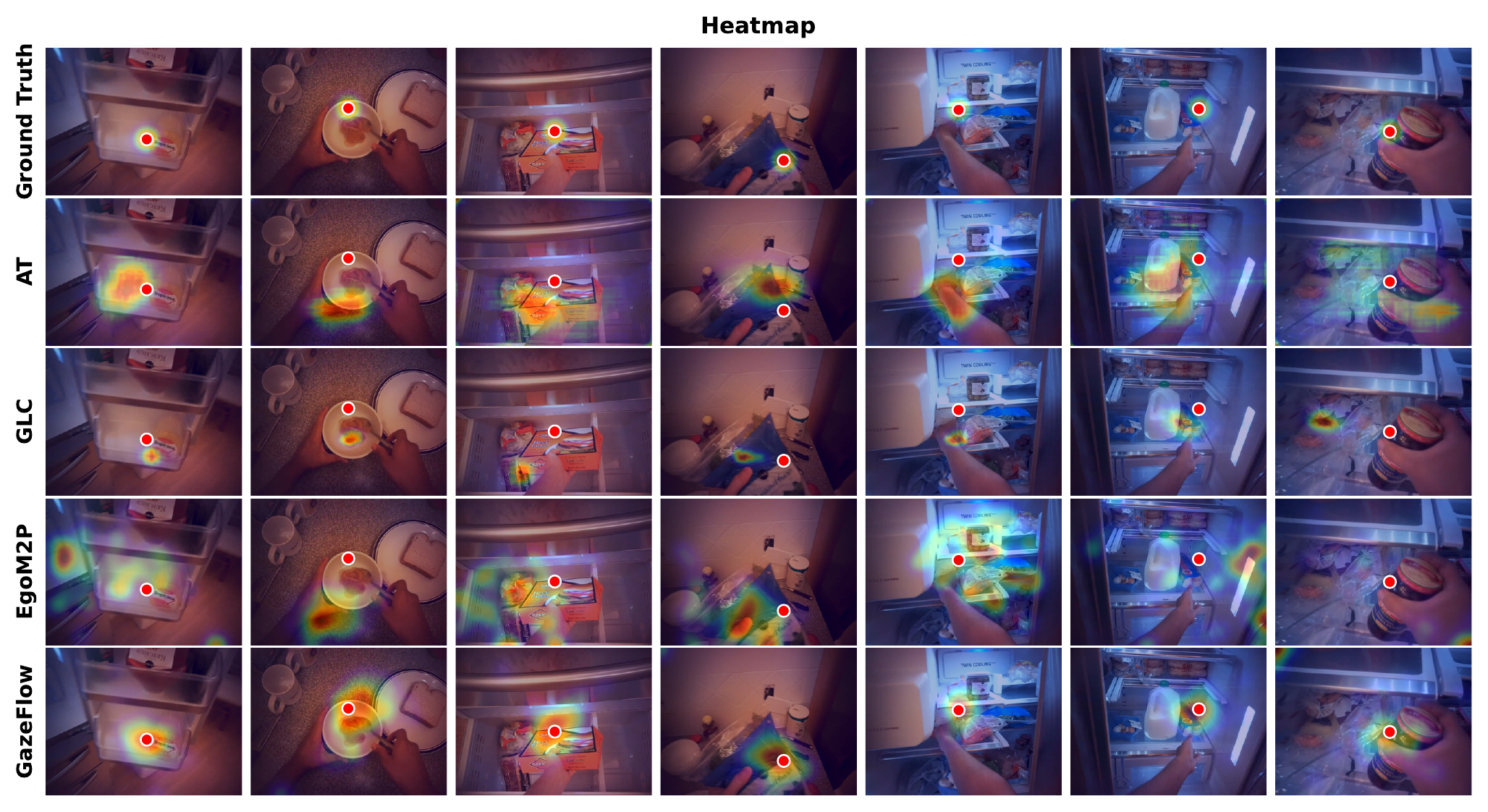}
  \caption{%
    \textbf{Qualitative comparison on EGTEA Gaze+.}
    Example frames where \proj predicts a heatmap tightly localized on the ground-truth gaze positions (red dots) compared to existing methods (AT~\citep{huang2018predicting}, GLC~\citep{lai2022glc}, EgoM2P~\citep{li2025egom2p}).
  }
  \label{fig:teaser}
  \vspace{-3pt}
\end{figure}

Taken together, gaze pattern is best modeled as a \textit{joint} distribution of temporal gaze positions conditioned upon the \textit{entire} video sequence.
Any gaze trajectory arising from interacting with the scene captured in the video can then be viewed as a sample from this distribution.
Modeling the joint distribution enables us to capture the global temporal dynamics of gaze, while conditioning on the full video allows us to account for both top-down influences, which arise from longer-term contextual and task-related information (on the order of seconds), and bottom-up saliency (e.g., local contrast or flicker), which operate at shorter timescales (tens of milliseconds).

We formalize these observations into a computational framework, \proj.
In \proj, egocentric gaze prediction is formulated as a generative process, where each generated trajectory is a sample from the underlying human gaze distribution.
In particular, we use conditional flow matching (CFM), a generative modeling method.
We use a video encoder to extract bottom-up spatio-temporal features from video frames.
To incorporate top-down influences, the encoded video embeddings are allowed to be globally queried by gaze at any time, yielding task-relevant information.
The bottom-up and top-down embeddings are then used to condition the generation process.

We evaluate \proj on two common egocentric video datasets, EGTEA Gaze+~\citep{li2018eye} and Ego4D~\citep{grauman2022ego4d}.
Not only does \proj achieve state-of-the-art performance on per-frame metrics, the predicted trajectories also exhibit closer alignment with the temporal dynamics of human gaze.
\Fig{fig:teaser} shows the qualitative results.
In summary, this paper makes the following contributions:
\begin{itemize}[leftmargin=1.2em, labelsep=0.4em, topsep=0pt, partopsep=0pt, itemsep=1pt, parsep=0pt]
    \item Starting from human gaze mechanisms, we first formulate egocentric gaze prediction as a conditional joint-distribution modeling problem.
    \item We propose a flow matching-based generative algorithm, where gaze trajectory generation is conditioned on both bottom-up (saliency) and top-down (task-specific) information from a video.
    \item We demonstrate that \proj achieves state-of-the-art per-frame prediction accuracy on EGTEA Gaze+ and Ego4D, and is better aligned with the temporal dynamics of human gaze.
\end{itemize}

\section{Related Work}
\label{sec:related}

\paragraph{Mechanisms of Human Gaze.}
Human gaze is shaped by three complementary mechanisms.
Bottom-up saliency is driven by low-level visual features such as contrast, orientation, and motion~\citep{itti1998saliency, itti2000saliency}.
Top-down task control guides the eyes to anticipate upcoming actions and select objects relevant to the current goal~\citep{yarbus1967eye,land1999roles,hayhoe2005eye,tatler2011eye, rothkopf2007task, zhao2026multidimensional}.
Oculomotor control further imposes structured temporal dynamics on gaze.
These dynamics alternate between saccades, fixations, and microsaccades, constraining how gaze evolves over time~\citep{rucci2018temporal}.
\proj{} addresses these factors jointly: a visual encoder supplies bottom-up evidence, a task-token bank supplies top-down context, and conditional flow matching captures temporal structure.

\paragraph{Joint Distribution Modeling.}
Modeling a high-dimensional joint distribution generally falls into two families.
Autoregressive factorization decomposes the joint into per-step conditionals~\citep{vaswani2017attention,brown2020language}, while distribution mapping transports a simple prior to the data via a learned map, as in GANs~\citep{goodfellow2014generative}, VAEs~\citep{kingma2014autoencoding}, diffusion~\citep{ho2020denoising}, and flow matching~\citep{lipman2023flow,tong2024improving}.
\proj{} adopts the latter, sampling a full trajectory through one video-conditioned Ordinary Differential Equation (ODE) transport (\Sect{sec:experiments}).

\paragraph{Egocentric Gaze Prediction.}
Existing methods fall into four paradigms summarized in Table~\ref{tab:paradigm_comparison}.

\emph{Regression} directly regresses one gaze point per frame from the video, trained with per-frame Mean Squared Error (MSE).
Early egocentric methods regress over hand-crafted cues~\citep{li2013learning}.
This collapses gaze to a single mode and cannot capture its multimodality.
\emph{Marginal density estimation} conditions on the video and independently predicts a per-frame heatmap per frame.
GLC~\citep{lai2022glc} produces this heatmap by correlating each frame with a global clip summary through a transformer.
CSTS~\citep{lai2024csts} additionally exploits sound, and DFG~\citep{Zhang_2017_CVPR} learns a GAN.
Per-frame factorization misses temporal coupling across frames.

\emph{Autoregression} factors the joint distribution into a chain of per-frame conditionals on previously generated gaze.
Attention Transition~\citep{huang2018predicting} predicts each fixation by recursively conditioning on the previous-frame fixation through an LSTM-based attention-transition module.
EgoM2P~\citep{li2025egom2p} decodes VQ-VAE-quantized gaze tokens within a multimodal pretraining stack.
This captures temporal dependency but accumulates errors under self-rollout~\citep{zhao2026reference}.
\emph{Distribution mapping} models the full trajectory directly, sampling each trajectory by transporting a simple prior through a learned map.
\proj{} adopts conditional flow matching~\citep{lipman2023flow,tong2024improving}: a learned velocity network defines a video-conditioned velocity field that transports each Gaussian noise sample into a gaze trajectory.
This jointly captures temporal coupling, stochasticity, and the bottom-up/top-down factors of natural gaze (\Sect{sec:gazeflow}).

\begin{table}[t]
  \centering
  \caption{\textbf{Comparison of egocentric gaze prediction paradigms.}
  $\mathcal{V}$ is the input video and $\mathcal{V}_{\le t}$ its causal prefix, $\mathbf{g}_t \in \mathbb{R}^2$ is the gaze position at frame $t$, $\mathbf{g}_{1:T}$ the full trajectory, and $q_\theta(\,\cdot\,\mid\,\cdot\,)$ denotes a model's parametric conditional distribution.
  MSE: mean squared error, KL: Kullback--Leibler divergence, and BCE: binary cross-entropy.
  $\checkmark$ if satisfied, $\times$ if not satisfied, and $\sim$ if architecturally implied but not enforced.
  }
  \label{tab:paradigm_comparison}
  \scriptsize
  \setlength{\tabcolsep}{3pt}
  \renewcommand{\arraystretch}{1.15}
  \begin{tabular}{@{}l l l l cccc@{}}
    \toprule
     & & & & \multicolumn{4}{c}{\textbf{Human gaze mechanism}} \\
    \cmidrule(lr){5-8}
    Method & Paradigm & Prediction & Training objective
      & Bottom-up & Top-down & Joint & Stochastic \\
    \midrule
    ~\citet{li2013learning}
      & Regression
      & $\mathbf{g}_t \mid \mathcal{V}$
      & per-frame MSE
      & $\times$ & $\sim$ & $\times$ & $\times$ \\
    \midrule
    GLC~\citep{lai2022glc}
      & \multirow{2}{*}{Marginal density estimation}
      & $q_\theta(\mathbf{g}_t \mid \mathcal{V})$
      & per-frame KL
      & $\checkmark$ & $\times$ & $\times$ & $\checkmark$ \\
    DFG~\citep{Zhang_2017_CVPR}
      &
      & $q_\theta(\mathbf{g}_t \mid \mathcal{V})$
      & per-frame KL
      & $\checkmark$ & $\times$ & $\sim$ & $\checkmark$ \\
    \midrule
    AT~\citep{huang2018predicting}
      & \multirow{2}{*}{Autoregressive}
      & $q_\theta(\mathbf{g}_t \mid \mathbf{g}_{t-1},\, \mathcal{V}_{\le t})$
      & per-frame BCE
      & $\checkmark$ & $\sim$ & $\sim$ & $\checkmark$ \\
    EgoM2P~\citep{li2025egom2p}
      &
      & $q_\theta(\mathbf{g}_t \mid \mathbf{g}_{<t},\, \mathcal{V})$
      & masked $L^2$
      & $\sim$ & $\sim$ & $\sim$ & $\checkmark$ \\
    \midrule
    \textbf{\proj{} (Ours)}
      & Distribution mapping
      & $q_\theta(\mathbf{g}_{1:T} \mid \mathcal{V})$
      & $L^2$ velocity field
      & $\checkmark$ & $\checkmark$ & $\checkmark$ & $\checkmark$ \\
    \bottomrule
  \end{tabular}
\end{table}

\section{\proj}
\label{sec:gazeflow}

In egocentric gaze prediction task, a head-mounted video $\mathcal{V}$ of $T$ frames and its ground-truth gaze trajectory $\mathbf{g}\in\mathbb{R}^{2T}$ are recorded jointly while human performs a task.
Gaze is inherently stochastic and depends on both what the human is doing and the visual saliency in the scene.
So $\mathbf{g}$ is one realization of a distribution conditioned on $\mathcal{V}$.
We write $p^\star(\mathbf{g}\mid\mathcal{V})$ for the population distribution over all gaze trajectories given $\mathcal{V}$, and each annotated trajectory in the dataset is one sample drawn from $p^\star$.
Our goal is to learn a model distribution $q_\theta(\mathbf{g}\mid\mathcal{V})$ that approximates $p^\star$, the underlying distribution we want to recover.
Since we have only samples from $p^\star$, not a closed-form density, we adopt conditional flow matching (CFM)~\citep{lipman2023flow,tong2024improving}, whose training objective is a sample-based regression that fits a velocity field transporting Gaussian noise to $q_\theta$ (see details in \Apx{app:flow}).

\begin{figure}[t]
  \centering
  \includegraphics[width=\linewidth]{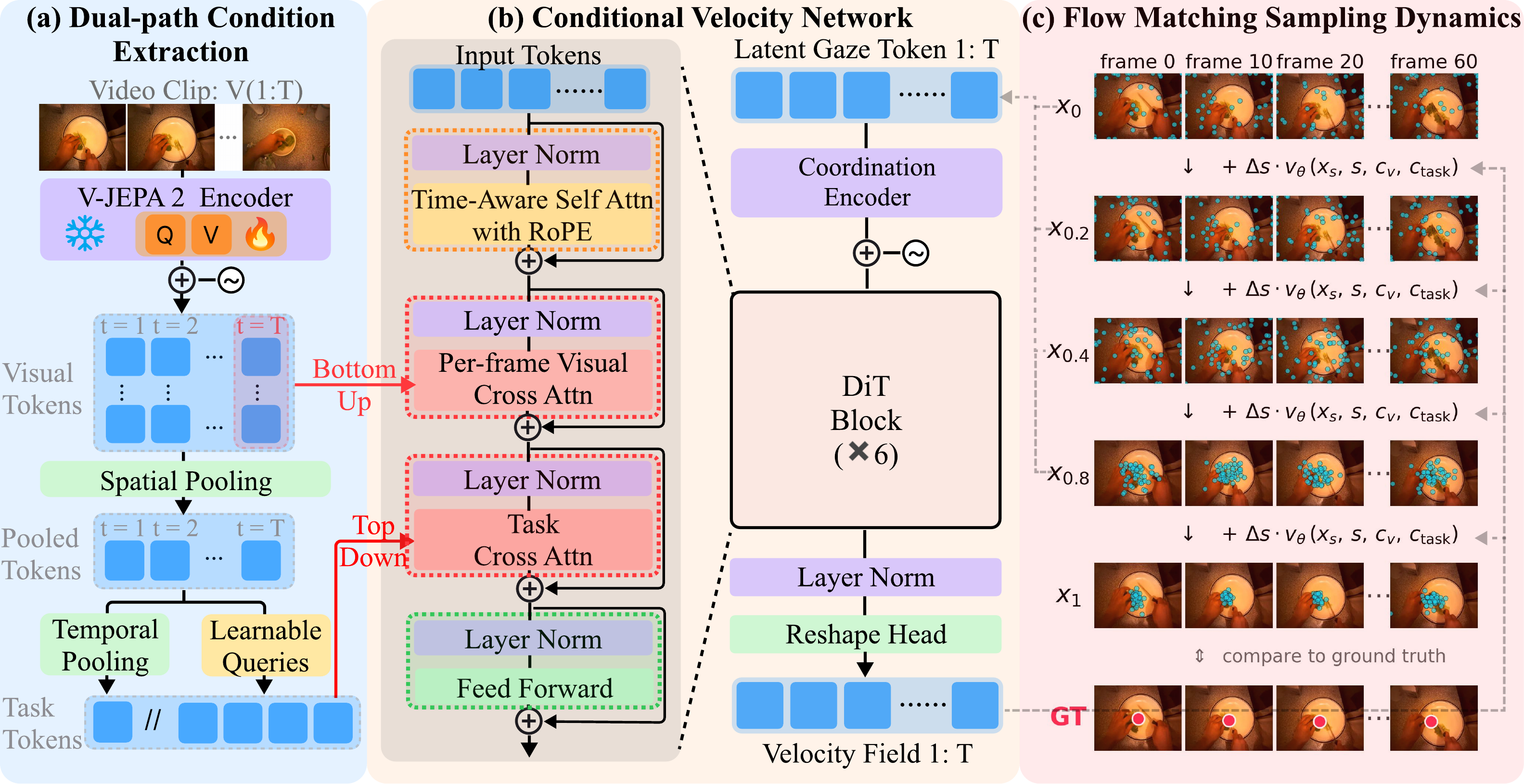}
  \caption{%
    \textbf{\proj architecture.}
    (a) Dual-path condition extraction (\Sect{sec:video_feature}).
    A LoRA-adapted V-JEPA~2 backbone produces bottom-up visual tokens $\mathbf{c}_v$; spatial--temporal pooling and learnable queries form the top-down task-token bank $\mathbf{c}_{\text{task}}$.
    (b) Conditional velocity network (\Sect{sec:method_cfm}).
    Latent gaze tokens encoding $\mathbf{x}_s$ and $s$ pass through six DiT blocks, each containing four sub-layers; a linear head outputs the velocity field $\mathbf{v}_\theta$.
    (c) Flow-matching sampling dynamics. 
    Euler integration of $\mathbf{v}_\theta$ transports Gaussian noise $\mathbf{x}_0$ to a clean trajectory $\mathbf{x}_1$; the cyan sample cloud at $s\!\in\!\{0,0.2,0.4,0.8,1\}$ contracts toward the ground truth.
  }
  \label{fig:architecture}
\end{figure}

\Fig{fig:architecture} shows the architecture of \proj, which contains two main components:
First, video feature extraction computes bottom-up visual tokens and top-down task tokens (\Sect{sec:video_feature}).
Second, the conditional velocity network uses these tokens for calculating a velocity transport that maps a Gaussian noise to target trajectory distribution (\Sect{sec:method_cfm}).
We discuss how our framework can deal with training and inference on long videos (\Sect{sec:training_inference}).

\subsection{Dual-Pathway Video Condition Extraction}
\label{sec:video_feature}

The video-conditioning module $E_\psi(\cdot)$ maps an input clip to two conditions.
The first condition is bottom-up and frame-local.
It comes from a pretrained video encoder.
The second condition is top-down and clip-level.
It comes from a learned task-token bank.
This dual-pathway design follows the gaze mechanisms reviewed in \Sect{sec:related}:
Human gazing depends on both low-level visual saliency and high-level task goals.
Formally,
\begin{equation}
    \mathbf{c} := \big(
    \mathbf{c}_v,
    \mathbf{c}_{\text{task}}
    \big)
    =
    E_\psi(\mathcal{V}),
    \label{eq:method:video_conditioning}
\end{equation}
where $\mathbf{c}_v$ contains frame-level visual tokens and $\mathbf{c}_{\text{task}}$ contains task-level tokens.

\paragraph{Bottom-Up Feature Extraction.}
The bottom-up path provides visual saliency for gaze prediction.
We use V-JEPA~2 ViT-L~\citep{assran2025vjepa2} as the pretrained video encoder.
We resize each frame ($T$ in total) to size of $256{\times}256$.
With a patch size of $N=16{\times}16$, each frame produces a $16{\times}16$ grid of spatial tokens.
A trainable linear layer projects each token to a width $d=256$:
\begin{equation}
    \mathbf{c}_v
    =
    \mathbf{V}
    \in
    \mathbb{R}^{T\times N\times d},
    \qquad
    N=16\times16.
    \label{eq:method:visual_condition}
\end{equation}
Each gaze token $\mathbf{h}_t$ (see \Sect{sec:method_cfm}) is allowed to cross-attend to the spatial tokens from the same frame $t$.
This biases $\mathbf{c}_v$ toward frame-specific evidence.
Modeling the temporal gaze correlation is handled by the velocity network (\Sect{sec:method_cfm}).

\paragraph{Top-Down Task Token Bank.}
\label{sec:global_aggregation}

Frame-level visual tokens alone do not encode task-level intent over the whole prediction horizon.
We therefore aggregate the video tokens into a small bank of task tokens.
This module provides the top-down task-level signal in \proj.

We first average the video tokens over the spatial axis to obtain one summary $ \bar{\mathbf{V}}_t$ per frame.
We then average the frame summaries over time to obtain one global video token $\mathbf{p}$.
This single token is stable, but it has limited representational power and could miss local task information.
We therefore add a learnable-query mechanism based on multi-head cross-attention (MHCA)~\cite{vaswani2017attention}.
Specifically, we initialize $M$ learnable task queries $\mathbf{Q}_{\text{task}}$ attend to the $T$ frame summaries:
\begin{equation}
    \bar{\mathbf{V}}_t
    =
    \frac{1}{N}
    \sum_{n=1}^{N}
    \mathbf{V}_{t,n,:},
    \qquad
    \mathbf{p}
    =
    \frac{1}{T}
    \sum_{t=1}^{T}
    \bar{\mathbf{V}}_t,
    \qquad
    \mathbf{q}_{1:M}
    =
    \mathrm{MHCA}
    \big(
    \mathbf{Q}_{\text{task}},
    \bar{\mathbf{V}}
    \big).
    \label{eq:method:task_slots}
\end{equation}
Here, $n$ indexes the $N$ spatial tokens within frame $t$.
$\bar{\mathbf{V}}=[\bar{\mathbf{V}}_1;\ldots;\bar{\mathbf{V}}_T]$ stacks all the frame summaries.
The global token $\mathbf{p}$ captures coarse task context while the queries $\mathbf{q}_{1:M}$ allow for more fine-grained, selective extraction of the task information from the video.
The final task-token bank is
\begin{equation}
    \mathbf{c}_{\text{task}}
    =
    [\mathbf{p};\mathbf{q}_{1:M}]
    \in
    \mathbb{R}^{(M+1)\times d}.
    \label{eq:method:task_bank}
\end{equation}
We verify this design in three ways.
\Sect{sec:task_probe} shows that the task tokens $\mathbf{c}_{\text{task}}$ encode task-level semantics.
\Sect{sec:design} shows that the learnable queries $\mathbf{q}_{1:M}$ strengthen task encoding and improve prediction compared with a single global summary.
\Sect{sec:ablation} shows that task information improves gaze prediction.
Appendix~\ref{app:architecture} gives all the details.

\subsection{Conditional Velocity Network}
\label{sec:method_cfm}

We now describe how the bottom-up embeddings $\mathbf{c}_{\text{v}}$ and top-down embeddings $\mathbf{c}_{\text{task}}$ are used to help generate gaze trajectories.
In particular, we use conditional flow matching~\citep{lipman2023flow,tong2024improving}, where generation is a transport process.
A sample starts as a Gaussian noise $\mathbf{x}_0\in\mathbb{R}^{2T}$, where $T$ is the number of discrete points (frames) on the gaze trajectory we predict (each of the discrete point has two coordinates, hence $2T$ dimensions).
It then passes through noisy trajectory states $\mathbf{x}_s$, and eventually reaches a clean trajectory sample $\mathbf{x}_1$.
This transport is realized by the following differential equation:
\begin{equation}
    \mathbf{x}_0 \sim \mathcal{N}(\mathbf{0},\mathbf{I}),
    \quad
    \frac{d\mathbf{x}_s}{ds}
    =
    \mathbf{v}_\theta(\mathbf{x}_s,s,\mathbf{c}),
    \quad\hat{\mathbf{g}} = 
    \mathbf{x}_1 = \int_0^1 \mathbf{v}_\theta(\mathbf{x}_s,s,\mathbf{c}) \text{d}s + \mathbf{x}_0
     \sim q_\theta(\mathbf{g}\mid\mathcal{V}).
    \label{eq:bg:cfm}
\end{equation}

Conditioned on $\mathbf{c}_v$ and $\mathbf{c}_{\text{task}}$ (collectively referred as $\mathbf{c}$), the velocity network $\mathbf{v}_\theta$ decides how the current state should move at flow time $s$.
$\mathbf{v}_\theta(\cdot)$ takes the current state $\mathbf{x}_s$, the flow time $s$, and the conditions $\mathbf{c}$ as input.
$\mathbf{x}_s$ and $s$ are first embedded into $T$ gaze tokens,
each of which corresponds to the gaze at frame $t$.
The gaze tokens are then updated in each denoising step by four separate blocks, which are visualized in \Fig{fig:architecture}(b).

First, \textbf{time-aware self-attention} lets gaze tokens exchange information over time.
This is where temporal correlation of gaze positions in the trajectory can be captured.
We use 1D RoPE to encode relative frame offsets~\citep{su2024roformer}.
Second, \textbf{per-frame visual cross-attention} lets each gaze token read the spatial visual tokens of the same frame from $\mathbf{c}_v$.
This injects bottom-up visual saliency.
Third, \textbf{task cross-attention} lets all gaze tokens read the task-token bank $\mathbf{c}_{\text{task}}$.
This injects top-down, task-specific information.
Fourth, a \textbf{feed-forward} layer transform each token independently.
We implement the velocity network with DiT-style blocks~\citep{peebles2023dit}.
\Apx{app:architecture} gives the full design details.
Finally, a linear output head maps the latent gaze tokens to velocity in the trajectory space $\mathbb{R}^{2T}$.

At inference time, the integral in \Equ{eq:bg:cfm} is evaluated numerically with a set of Euler steps, each of which evaluates $\mathbf{v}_\theta(\cdot)$ once.
Each integral evaluation uses a noise sample $\mathbf{x}_0$ and gives a $\mathbf{x}_1$, representing one sample of the underlying distribution of the gaze trajectory $q_\theta(\mathbf{g}\mid\mathcal{V})$.
Repeating this integration with different noise samples gives a sample distribution of the gaze trajectory.

\subsection{Training \& Inference}
\label{sec:training_inference}

\paragraph{Training.}
We train \proj{} on 64-frame clips paired with per-frame gaze.
For each clip, we sample $s \sim \mathcal{U}[0,1]$ and $\mathbf{x}_0 \sim \mathcal{N}(\mathbf{0},\mathbf{I}_{2T})$, and form $\mathbf{x}_s = (1-s)\mathbf{x}_0 + s\mathbf{g}$ with a sample $\mathbf{g}\sim p^\star(\cdot\mid\mathcal{V})$ given by the dataset.
The CFM loss regresses $\mathbf{v}_\theta$ on the per-sample velocity $\mathbf{g}-\mathbf{x}_0$:
\begin{equation}
    \mathcal{L}_{\mathrm{CFM}}(\theta)
    \;=\;
    \mathbb{E}_{\mathcal{V},\,s,\,\mathbf{x}_0,\,\mathbf{g}}\!
    \Bigl[\,
    \bigl\|\,
    \mathbf{v}_\theta(\mathbf{x}_s,\, s,\, \mathbf{c})
    -
    (\mathbf{g} - \mathbf{x}_0)
    \,\bigr\|_2^2
    \,\Bigr].
    \label{eq:method:cfm_loss}
\end{equation}
See \Apx{app:alg} and \Apx{app:selfcond} for the full training algorithm and engineering details.

\paragraph{Long-Video Inference.}
We process videos longer than 64 frames with overlapping windows.
Each window is encoded once for $(\mathbf{c}_v, \mathbf{c}_{\text{task}})$.
We integrate \Eqn{eq:bg:cfm} via Euler steps in parallel with multiple noise samples; 
each endpoint $\mathbf{x}_1$ is one trajectory sample for that window.
For adjacent windows, we linearly blend predictions in the overlap region by frame index.
This keeps transitions continuous and avoids encoding the full video at once.
\Apx{app:window} gives the exact windowing and blending rule.

\section{Experiments}
\label{sec:experiments}

\subsection{Experimental Setup}
\label{sec:setup}

\paragraph{Training and Inference.}
We optimize both the video encoder and the velocity network jointly in an end-to-end manner.
The encoder is adapted via LoRA, applying rank-$16$ low-rank~\citep{hu2022lora} updates to its query and value projections, while the velocity network is trained with full parameters.
The model is trained by minimizing the conditional flow-matching objective $\mathcal{L}_{\mathrm{CFM}}$ in \Eqn{eq:method:cfm_loss} using AdamW~\citep{loshchilov2019decoupled} ($\beta_1{=}0.9$, $\beta_2{=}0.999$, weight decay $10^{-4}$), for $80$ epochs with a base learning rate of $10^{-4}$ under a cosine schedule decaying to $10^{-6}$.
All experiments are run on $4{\times}$NVIDIA~H100 GPUs in bfloat16 mixed precision.
Remaining hyperparameters and configurations are summarized in Appendix~\ref{app:setup_training}.
At inference, we draw $K{=}50$ trajectory samples per clip in parallel as a Monte Carlo of the learned joint distribution.
Each sample is generated by integrating the learned velocity field $\mathbf{v}_\theta$ using the explicit Euler scheme over $S{=}50$ uniform steps, starting from an independent draw of $\mathcal{N}(\mathbf{0},\mathbf{I})$.
It takes $\sim\!3.4$\,s to inference a clip of $T{=}64$ frames on a single H100 (\Apx{app:efficiency}).
With $K{=}30$ and $S{=}5$, inference is $4.6\times$ faster with no loss in accuracy (\Apx{app:ks_grid}).

\paragraph{Per-Frame Heatmaps.}
At each frame, we convert the $K{=}50$ sampled gaze points into a per-frame heatmap by placing a Gaussian kernel at each point and averaging.
This makes \proj directly comparable to heatmap-output baselines under the heatmap-based per-frame metrics defined below.

\paragraph{Datasets.}
We evaluate on two egocentric gaze datasets.
\underline{EGTEA Gaze+}~\citep{li2018eye} contains cooking videos recorded at $640\times480$ resolution and $24$\,fps, with eye-tracker gaze annotations, and provides $8{,}299$ training and $2{,}022$ validation clips.
\underline{Ego4D}~\citep{grauman2022ego4d} spans a broader range of daily activities; we use the subset with gaze annotations, recorded at $1088\times1080$ resolution and $30$\,fps, with $12{,}178$ training and $5{,}202$ validation clips.
Data preparation details are provided in Appendix~\ref{app:setup_datasets}.

\paragraph{Baselines.}
We compare against three families of prior methods. For the first two families, we include representative state-of-the-art models.
\begin{itemize}[leftmargin=1.2em, labelsep=0.4em, topsep=0pt, partopsep=0pt, itemsep=1pt, parsep=0pt]
\item \emph{Marginal density estimation.} \underline{GLC}~\citep{lai2022glc}, a state-of-the-art egocentric gaze model that outputs a per-frame fixation heatmap independently for each frame.
\item \emph{Autoregressive.} \underline{AT}~\citep{huang2018predicting}, the first egocentric gaze model to learn attention transitions using an LSTM conditioned on previous-frame gaze, and \underline{EgoM2P}~\citep{li2025egom2p}, a multimodal model pretrained on large-scale egocentric datasets that decodes gaze as a discrete autoregressive token sequence.
\item \emph{Simple Priors.} \underline{Center Bias} always predicts the gaze position as the image center; \underline{Random Walk} is a Gaussian random walk whose step variance is set to the human frame-to-frame gaze motion calculated from the dataset.
\end{itemize}

We retrain AT, use the official checkpoint for GLC, and evaluate EgoM2P in two settings: the official zero-shot checkpoint and LoRA fine-tuning.
All baselines are evaluated on both datasets, with full configuration details provided in Appendix~\ref{app:setup_baselines}.

\paragraph{Metrics.}
We use two commonly used metrics: per-frame prediction accuracy and alignment with human temporal dynamics.
The first asks whether gaze is predicted at the correct spatial location on each frame.
The second asks whether the predicted trajectory has a human-like temporal structure.
Full definitions, interpretations, and computation details of all metrics are provided in Appendix~\ref{app:setup_metrics}.

\textbf{(1) Per-Frame Accuracy.} We report the \emph{Area Under the Curve} (AUC)~\citep{bylinskii2019different, Judd_2012}, \emph{F1}~\citep{lai2022glc}, \emph{Precision} (P)~\citep{lai2022glc}, \emph{Recall} (R)~\citep{lai2022glc}, and \emph{Average Angular Error} (AAE)~\citep{huang2018predicting}.

\textbf{(2) Temporal Dynamics Alignment.} We compare each predicted trajectory against the human trajectory using the \emph{Average Displacement Error} (ADE)~\citep{kara2025diffeye} and \emph{Dynamic Time Warping distance} (DTW)~\citep{pellegrini2009you, kara2025diffeye}, each reported as mean and best-of-$K$ over $K{=}50$ samples.
Additionally, we report the mean and median \emph{per-frame displacement} as ratios to the human reference, and the \emph{Jensen--Shannon divergence} (JSD)~\citep{fuglede2004jensen} between the predicted and human displacement histograms.

\paragraph{Protocol.}
All per-frame evaluations follow the GLC-format protocol~\citep{lai2022glc}.
For each validation clip, we sample uniformly spaced frames, compute metrics against the ground-truth fixation maps, average over frames within the clip, and then average over clips.
The main comparisons in \Sect{sec:results} use the LoRA variant of \proj{} on the full validation split.
Ablations in \Sect{sec:ablation} and design analyses in \Sect{sec:design} use the frozen-encoder variant on a fixed validation subset, since repeating $K{=}50$ LoRA sampling per variant is computationally expensive.
Details on subset construction, frozen-encoder evaluation, and sample-count selection are provided in \Apx{app:setup_metrics} and \Apx{app:k_sample_sweep}.
\subsection{\proj Outperforms Prior Methods}
\label{sec:results}

Joint distribution modeling achieves the best per-frame accuracy across datasets.
\Tbl{tab:main} shows that \proj outperforms all baselines on both EGTEA Gaze+ and Ego4D.
Compared to the strongest baseline, \proj{} improves F1 by $16.6\%$ and $10.4\%$ on the two datasets, reflecting better binarized fixation prediction under the GLC evaluation protocol.
It also reduces AAE by $8.8\%$ and $9.2\%$ on the two datasets, corresponding to lower angular error of the predicted fixation with respect to human gaze.
\Tbl{tab:main_full} shows the full breakdown.

\begin{table}[t]
  \centering
  \caption{\textbf{Per-frame accuracy on EGTEA Gaze+ and Ego4D.}
  \proj achieves the best performance; results on more metrics are in \Tbl{tab:main_full} in \Apx{app:full_main_table}.
  Best per column is in bold.
  }
  \label{tab:main}
  \scriptsize
  \setlength{\tabcolsep}{2pt}
  \renewcommand{\arraystretch}{1}
  \begin{tabular}{@{}l l ccccc ccccc@{}}
    \toprule
    & & \multicolumn{5}{c}{\textbf{EGTEA Gaze+}} & \multicolumn{5}{c}{\textbf{Ego4D}} \\
    \cmidrule(lr){3-7} \cmidrule(lr){8-12}
    Paradigm & Method
      & AUC$\uparrow$ & F1$\uparrow$ & P$\uparrow$ & R$\uparrow$ & AAE$\downarrow$
      & AUC$\uparrow$ & F1$\uparrow$ & P$\uparrow$ & R$\uparrow$ & AAE$\downarrow$ \\
    \midrule
    \multirow{2}{*}{\emph{Simple Priors}}
      & Center Bias                                      & 0.906 & 0.185 & 0.162 & 0.215 & 16.07 & 0.930   & 0.222   & 0.194   & 0.258   & 13.92   \\
      & Random Walk                                      & 0.824 & 0.147 & 0.090 & 0.413 & 16.34 & 0.763   & 0.109   & 0.059   & \textbf{0.662}   & 14.38   \\
    \cmidrule(lr){1-12}
    \emph{Marginal density}
      & GLC~\citep{lai2022glc}                           & 0.953 & 0.421 & 0.348 & 0.531 & 10.21 & 0.947   & 0.355   & 0.267   & 0.530   & 11.52   \\
    \cmidrule(lr){1-12}
    \multirow{3}{*}{\emph{Autoregressive}}
      & EgoM2P (zero-shot)~\citep{li2025egom2p}          & 0.921 & 0.280 & 0.195 & 0.498 & 13.18 & 0.911 & 0.253 & 0.180 & 0.422 & 13.86 \\
      & EgoM2P (LoRA)~\citep{li2025egom2p}               & 0.921 & 0.293 & 0.205 & 0.515 & 12.81 & 0.930 & 0.297 & 0.217 & 0.470 & 12.69 \\
    & AT~\citep{huang2018predicting}                   & 0.956 & 0.419 & 0.339 & 0.547 & 9.88  & 0.954   & 0.346   & 0.268   & 0.487   & 12.01   \\
    \cmidrule(lr){1-12}
    \emph{Distribution Mapping}
      & \textbf{\proj{} (ours)}                          & \textbf{0.964} & \textbf{0.491} & \textbf{0.414} & \textbf{0.603} & \textbf{9.01} & \textbf{0.958} & \textbf{0.392} & \textbf{0.307} & 0.540 & \textbf{10.46} \\
    \bottomrule
  \end{tabular}
\end{table}

\proj also matches human gaze dynamics much more closely than prior methods.
\Tbl{tab:dynamics} reports the alignment-with-human-dynamics metrics.
\proj achieves the lowest mean and best-of-$K$ ADE and DTW, indicating smaller path-wise error to the ground-truth trajectory under Euclidean and temporally aligned comparisons, respectively.
On best-of-$K$, \proj reduces this error by roughly $1.6\times$ compared to the strongest baseline, EgoM2P with LoRA, with ADE decreasing from $0.96$ to $0.60$ and DTW from $0.81$ to $0.48$.
Our per-step displacement statistics are also closest to the human reference: the median is $1.08\times$ human and the mean follows the same pattern, whereas GLC and AT are $11.75\times$ and $24.55\times$, respectively.
The displacement JSD is $0.002$, the lowest among all methods and closest to the human reference distribution.

\begin{table}[t]
  \caption{\textbf{Alignment with human gaze dynamics on EGTEA Gaze+.} Disp Mean and Disp Med are per-step gaze displacements shown as ratios to the human trajectory (subheader \texttt{H}, measured in pixels; closer to $1\times$ is better). Best per column in bold.}
  \label{tab:dynamics}
  \centering
  \scriptsize
  \setlength{\tabcolsep}{4pt}
  \renewcommand{\arraystretch}{0.95}
  \begin{tabular}{@{}l l cc cc cc c@{}}
    \toprule
    & & \multicolumn{2}{c}{ADE ($\times 10^{2}$\,px)$\downarrow$} & \multicolumn{2}{c}{DTW ($\times 10^{2}$\,px)$\downarrow$}
      & Disp Mean & Disp Med
      & JSD$\downarrow$ \\
    \cmidrule(lr){3-4}\cmidrule(lr){5-6}
    Paradigm & Method
      & Mean & Best & Mean & Best
      & (\texttt{H}: $12.5$\,px) & (\texttt{H}: $4.0$\,px)
      & \\
    \midrule
    \multirow{2}{*}{\emph{Simple Priors}}
      & Center Bias                                      & 1.35 & 1.35 & 1.35 & 1.35 & $0.00\times$ & $0.00\times$ & $0.125$ \\
      & Random Walk                                      & 2.07 & 1.05 & 1.97 & 0.92 & $2.54\times$ & $7.43\times$ & $0.294$ \\
    \cmidrule(lr){1-9}
    \emph{Marginal density}
      & GLC~\citep{lai2022glc}                           & 1.13 & 1.01 & 1.03 & 0.87 & $5.77\times$ & $11.75\times$ & $0.352$ \\
    \cmidrule(lr){1-9}
    \multirow{3}{*}{\emph{Autoregressive}}
      & EgoM2P (zero-shot)~\citep{li2025egom2p}          & 1.47 & 0.97 & 1.35 & 0.83 & $2.20\times$ & $3.05\times$ & $0.058$ \\
      & EgoM2P (LoRA)~\citep{li2025egom2p}               & 1.50 & 0.96 & 1.37 & 0.81 & $2.69\times$ & $3.03\times$ & $0.058$ \\
      & AT~\citep{huang2018predicting}                   & 1.32 & 1.12 & 1.22 & 0.99 & $9.70\times$ & $24.55\times$ & $0.476$ \\
    \cmidrule(lr){1-9}
    \emph{Distribution Mapping}
      & \textbf{\proj{} (ours)}                          & \textbf{1.04} & \textbf{0.60} & \textbf{0.95} & \textbf{0.48}
                                                        & $\mathbf{0.82\times}$ & $\mathbf{1.08\times}$
                                                        & $\mathbf{0.002}$ \\
    \bottomrule
  \end{tabular}
\end{table}

\subsection{Task Tokens Encode Task Semantics}
\label{sec:task_probe}

We have hypothesized that task information can be beneficial to gaze prediction, since human gazing is task dependent.
This hypothesis motivates us to extract the task-dependent tokens $\mathbf{c}_{\text{task}}$, which are used to condition the flow matching.
We now show that $\mathbf{c}_{\text{task}}$ does carry task-level semantics.

Each video clip in EGTEA Gaze+ is annotated at three label hierarchies: $19$ verbs (e.g., ``cut''), $53$ nouns (e.g., ``tomato''), and $106$ actions formed as verb-noun pairs (e.g., ``cut tomato'').
We freeze \proj, extract its task-token bank for each clip, and train a linear classifier to predict the action, verb, and noun labels.
We report Top-$k$ accuracy, which denotes the percentage of validation clips for which the ground-truth class appears among the linear classifier's $k$ highest-scoring predictions.
As shown in~\Tbl{tab:probing}, the global token $\mathbf{p}$ is already strongly predictive, reaching $49.1\%$, $68.3\%$, and $58.7\%$ Top-$1$ accuracy on action, verb, and noun, respectively.
The learnable task queries $\mathbf{q}_{1:M}$ are weaker in isolation, but provide complementary information: combining $\mathbf{p}$ and $\mathbf{q}_{1:M}$ improves Top-1 by $+5.5$, $+7.7$, and $+5.7$ points.
These results show that task-relevant semantics emerge in the task-token bank despite the absence of explicit supervision, consistent with the design in \Eqn{eq:method:task_bank}.

\begin{table}[t]
  \begin{minipage}[t]{0.37\linewidth}
    \centering
    \scriptsize
    \setlength{\tabcolsep}{2pt}
    \renewcommand{\arraystretch}{0.95}
    \caption{\textbf{Linear probing of the task-token bank} on EGTEA action/verb/noun labels. The $\cup$ indicates a clip is counted correct if at least one probe is correct.
    }
    \label{tab:probing}
    \begin{tabular}{@{}l cc cc cc@{}}
      \toprule
      & \multicolumn{2}{c}{Action (106)} & \multicolumn{2}{c}{Verb (19)} & \multicolumn{2}{c}{Noun (53)} \\
      \cmidrule(lr){2-3} \cmidrule(lr){4-5} \cmidrule(lr){6-7}
      Probe input & T1 & T5 & T1 & T5 & T1 & T5 \\
      \midrule
      $\mathbf{q}_{1:M}$ alone               & 40.5 & 66.6 & 64.1 & 95.9 & 47.9 & 78.7 \\
      $\mathbf{p}$ alone                     & 49.1 & 76.6 & 68.3 & 98.2 & 58.7 & 87.3 \\
      \textbf{$\mathbf{p} \cup \mathbf{q}_{1:M}$}
                                  & \textbf{54.6} & \textbf{80.2}
                                  & \textbf{76.0} & \textbf{98.7}
                                  & \textbf{64.4} & \textbf{89.8} \\
      \bottomrule
    \end{tabular}
  \end{minipage}\hfill
  \begin{minipage}[t]{0.61\linewidth}
    \centering
    \scriptsize
    \setlength{\tabcolsep}{3pt}
    \renewcommand{\arraystretch}{1.1}
    \caption{\textbf{Design alternatives to \proj{}.}
      \emph{w/o Learnable Queries}: drop the $M{=}4$ learnable queries (keep only the global mean-pool token).
      \emph{Concat}: replace task cross-attention with channel-wise concatenation.
      \emph{Learnable PE}: replace RoPE with a learnable absolute position embedding.}
    \label{tab:pe_compare}
    \begin{tabular}{@{}l ccccc@{}}
      \toprule
      Variant & AUC$\uparrow$ & F1$\uparrow$ & P$\uparrow$ & R$\uparrow$ & AAE$\downarrow$ \\
      \midrule
      \textbf{Default}      & \textbf{0.970} & \textbf{0.493} & \textbf{0.401} & \textbf{0.638} & \textbf{8.81} \\
      w/o Learnable Queries & 0.967          & 0.484          & 0.395          & 0.626          & 8.93 \\
      Concat                & 0.966          & 0.469          & 0.379          & 0.615          & 9.10 \\
      Learnable PE          & 0.967          & 0.476          & 0.380          & 0.635          & 8.94 \\
      \bottomrule
    \end{tabular}
  \end{minipage}
\end{table}

\subsection{Ablation Studies}
\label{sec:ablation}

We isolate the contribution of each architectural component by removing one component at a time, as shown in \Tbl{tab:ablation}.
The four ablation rows test the four architectural choices made in \Sect{sec:gazeflow}: joint-distribution modeling via flow matching, the bottom-up pathway, the top-down pathway, and 1D temporal RoPE for time-aware self-attention (SA).

\paragraph{Flow Matching vs. Regression.}
Replacing the flow-matching head with a deterministic regression head over the same encoder and conditioning inputs reduces AUC from $0.970$ to $0.903$ and inflates the displacement JSD against human gaze by roughly $66\times$.
This is the most direct evidence for the central claim of \Sect{sec:method_cfm}: gaze should be modeled as a joint distribution over the full trajectory.
A regression head, which neither models a distribution nor couples frames, predicts each gaze point in isolation and therefore fails to capture the temporal coupling between frames.

\paragraph{Flow Matching vs. Diffusion.}
We also replace the flow-matching objective with a diffusion objective, keeping the architecture, conditioning, and training setup the same.
At $50$ sampling steps, flow matching reaches F1 $0.493$ and diffusion $0.485$.
The gap grows as the step count drops: at $5$ steps, flow matching reaches F1 $0.506$ while diffusion drops to $0.359$.
Flow matching thus needs fewer steps and less inference time for the same accuracy (\Apx{app:fm_vs_diffusion}).

\paragraph{Dual-Path Conditioning.}
Dropping the per-frame visual cross-attention sub-layer of the velocity network, which carries the bottom-up condition $\mathbf{c}_v$ from \Sect{sec:video_feature}, lowers F1 from $0.493$ to $0.431$.
Dropping the task cross-attention sub-layer, which carries the top-down condition $\mathbf{c}_{\text{task}}$, lowers F1 to $0.476$.
Both pathways are individually necessary, supporting the dual-pathway design in \Sect{sec:video_feature}.

\paragraph{Temporal RoPE.}
Removing the relative-time RoPE from the time-aware self-attention sub-layer of the velocity network has little effect on per-frame accuracy but weakens the displacement JSD by roughly $2.5\times$ and worsens the median displacement to $1.4\times$ of that of human.
This confirms that the relative-offset encoding is what enables time-aware self-attention to capture inter-frame temporal coupling, as claimed in \Sect{sec:method_cfm}.

\begin{table}[t]
  \caption{\textbf{Ablation study.} \textbf{Full} is the default \proj{}; each remaining row ablates one architectural component while keeping all others identical, removing either the bottom-up visual feature ($\mathbf{c}_v$, frame-local visual cross-attention), the top-down task guidance ($\mathbf{c}_{\text{task}}$, task cross-attention), the relative-time RoPE in time-aware self-attention, or the joint-distribution flow-matching head (replaced with a deterministic regression head over the same conditioning).
  Per-column best in \textbf{bold}.}
  \label{tab:ablation}
  \centering
  \scriptsize
  \setlength{\tabcolsep}{4pt}
  \renewcommand{\arraystretch}{0.95}
  \begin{tabular}{@{}l ccccc cc ccc@{}}
    \toprule
    Variant
      & AUC$\uparrow$ & F1$\uparrow$ & P$\uparrow$ & R$\uparrow$ & AAE$\downarrow$
      & ADE$\downarrow$ & DTW$\downarrow$
      & Disp Mean & Disp Med & JSD$\downarrow$ \\
    & & & & &
      & &
      & (H: $12.5$\,px) & (H: $4.0$\,px) & \\
    \midrule
    \textbf{Full}
      & \textbf{0.970} & \textbf{0.493} & \textbf{0.401} & \textbf{0.638} & \textbf{8.81}
      & \textbf{0.57} & \textbf{0.45}
      & $\mathbf{0.91\times}$ & $\mathbf{1.08\times}$ & \textbf{0.0011} \\
    - Bottom-up Visual Feature
      & 0.961 & 0.431 & 0.333 & 0.613 & 9.22
      & 0.65 & 0.53
      & $0.89\times$ & $1.17\times$ & 0.0012 \\
    - Top-down Task Guidance
      & 0.965 & 0.476 & 0.380 & 0.638 & 9.14
      & 0.58 & 0.46
      & $0.92\times$ & $1.09\times$ & 0.0013 \\
    - Temporal RoPE
      & 0.968 & 0.477 & 0.388 & 0.620 & 8.94
      & 0.59 & 0.46
      & $1.21\times$ & $1.40\times$ & 0.0027 \\
    - Joint Distribution Modeling
      & 0.903 & 0.451 & 0.394 & 0.526 & 9.12
      & 0.84 & 0.79
      & $0.71\times$ & $2.01\times$ & 0.0729 \\
    \bottomrule
  \end{tabular}
\end{table}

\subsection{Design Choice Analysis}
\label{sec:design}
\label{sec:attention_analysis}

\paragraph{Temporal Gaze Correlation.}
Time-aware gaze-token self-attention is where our denoise block models the temporal coupling needed for joint-distribution modeling, and its temporal positional encoding decides how each gaze token reasons about its position relative to other frames.
We compare two choices: \emph{(i)} RoPE (default, details in \Apx{app:rope}), which injects each pairwise frame offset directly into the attention dot product, and \emph{(ii)} a learnable absolute position embedding, which assigns a learned vector to each frame index.
Replacing RoPE with the absolute embedding lowers F1 from $0.493$ to $0.476$ (\Tbl{tab:pe_compare}).
This is because gaze dynamics are properties of how far apart two frames are, not of where a frame sits inside an arbitrary clipped window.

\paragraph{Top-Down Task Guidance.}
Top-down task information from the task-token bank is injected into gaze tokens via a per-block task cross-attention sub-layer.
We compare two forms of task guidance: \emph{(i)} cross-attention (default; \Apx{app:task-cross-atten}), where each gaze token queries the bank at every denoiser block, and \emph{(ii)} channel-wise concatenation of the bank to the gaze-token input.
Concatenation degrades all metrics, including F1 ($0.493 \to 0.469$) and AAE ($8.81 \to 9.10$); see \Tbl{tab:pe_compare} for details.

Removing the $M{=}4$ learnable queries, leaving only the global mean-pool token, also reduces F1 to $0.484$ (\Tbl{tab:pe_compare}), confirming that the queries enable task-aware selective extraction beyond the low-bandwidth global token alone.

\paragraph{Task-Token Bank Bandwidth.}
We sweep $M\in\{1,2,4,8,16\}$, the number of learnable task queries, and the results are shown in \Tbl{tab:k_sweep} in \Apx{app:k_sweep}.
The performance peaks at $M=4$, indicating that too few queries cannot represent the diversity of task-relevant factors, while too many dilute the per-query specialization that cross-attention relies on.

\subsection{Interpretations}
\label{subsec: interpretation}
\paragraph{Interpreting Self-Attention.}
Self-attention in our velocity network decides, when predicting gaze at frame $i$, how strongly to draw on gaze at each other frame $j$.
For each head we measure how attention mass distributes over $j$ and find two regimes: \emph{look-back} heads concentrate on past frames and \emph{look-ahead} heads concentrate on future frames.
This pattern parallels two well-known modes of human gaze in egocentric tasks --- anticipating upcoming events~\citep{mennie2007lookahead,land2001roles} and lagging ongoing ones to verify completion~\citep{ballard1995memory,hayhoe2005eye}.
A third \emph{split-attention} category and full visualizations are in \Apx{app:head-attention}.

\paragraph{Interpreting Bottom-Up Saliency.}
The visual cross-attention sub-layer provides bottom-up guidance that decides, when predicting gaze at frame $t$, which spatial regions of the video to focus on.
Using off-the-shelf hand and active-object segmentation algorithm~\citep{liu2024grounding}, we measure how much attention mass each attention head places on the hand, the active object, and the rest of the scene.
We find that the heads cluster into hand-focused, object-focused, and coupled groups, and the cluster also varies with the action being performed.
This finding is consistent with the manipulation literature, in which gaze concentrates on the hands and the objects they act on~\citep{land1999roles,hayhoe2005eye}.
Details are in \Apx{app:head-semantics}.

\subsection{Limitations}
\label{subsec:limitation}
\proj's failures concentrate on clips with saccades whose endpoint lies outside the camera's field of view: the saccade target was never captured by the recorded video on which the model conditions, so there is no visual evidence for us to make the prediction (\Apx{app:failure}).

\proj models the conditional distribution of gaze trajectories given a fixed input video, so the head movement that produced that video is part of the conditioning.
The gaze distribution we recover is therefore of the residual eye movement, not the full distribution that would arise across observers freely interacting with the same scene, where head movement itself varies across trials~\citep{zhao2025coordauth}.
We leave to future work to jointly model head and eye movements. 

Additionally, \proj's prediction is also conditioned upon the \textit{entire} video sequence, which limits its applicability to scenarios that require real-time gaze prediction from egocentric videos such as gaze-contingent rendering~\cite{lin2025metasapiens, chen2025modeling, duinkharjav2022color, lin2025powergs}, where one has no access to future frames. 
We also train a causal variant that sees only past and current frames; it enables online gaze prediction at some cost in accuracy (details in \Apx{app:causal}).

\section{Conclusion}
\label{sec:conclusion}

\proj is designed to account for the three factors that influence the stochasticity of human gaze: structured temporal dynamics, top-down task influence, and bottom-up scene saliency.
\proj sets a new state of the art on per-frame prediction accuracy while simultaneously achieving the closest alignment with human gaze dynamics.
We hope \proj can broaden the use of gaze in applications that have traditionally required dedicated eye-tracking hardware.

\bibliography{references}

@inproceedings{leng2019energy,
  title={Energy-efficient video processing for virtual reality},
  author={Leng, Yue and Chen, Chi-Chun and Sun, Qiuyue and Huang, Jian and Zhu, Yuhao},
  booktitle={Proceedings of the 46th International Symposium on Computer Architecture},
  pages={91--103},
  year={2019}
}

@article{chen2025modeling,
  title={Modeling and Exploiting the Time Course of Chromatic Adaptation for Display Power Optimizations in Virtual Reality},
  author={Chen, Ethan and Kondguli, Sushant and Marshall, Carl and Zhu, Yuhao},
  journal={ACM Transactions on Graphics (TOG)},
  volume={44},
  number={6},
  pages={1--21},
  year={2025},
  publisher={ACM New York, NY, USA}
}

@article{duinkharjav2022color,
  title={Color-perception-guided display power reduction for virtual reality},
  author={Duinkharjav, Budmonde and Chen, Kenneth and Tyagi, Abhishek and He, Jiayi and Zhu, Yuhao and Sun, Qi},
  journal={ACM Transactions on Graphics (TOG)},
  volume={41},
  number={6},
  pages={1--16},
  year={2022},
  publisher={ACM New York, NY, USA}
}

@inproceedings{lin2025metasapiens,
  title={Metasapiens: Real-time neural rendering with efficiency-aware pruning and accelerated foveated rendering},
  author={Lin, Weikai and Feng, Yu and Zhu, Yuhao},
  booktitle={Proceedings of the 30th ACM International Conference on Architectural Support for Programming Languages and Operating Systems, Volume 1},
  pages={669--682},
  year={2025}
}

@inproceedings{lin2025powergs,
  title={PowerGS: Display-Rendering Power Co-Optimization for Neural Rendering in Power-Constrained XR Systems},
  author={Lin, Weikai and Kondguli, Sushant and Marshall, Carl and Zhu, Yuhao},
  booktitle={Proceedings of the SIGGRAPH Asia 2025 Conference Papers},
  pages={1--12},
  year={2025}
}

@inproceedings{geisler1998real,
  title={Real-time foveated multiresolution system for low-bandwidth video communication},
  author={Geisler, Wilson S and Perry, Jeffrey S},
  booktitle={Human vision and electronic imaging III},
  volume={3299},
  pages={294--305},
  year={1998},
  organization={SPIE}
}

@inproceedings{mazumdar2019perceptual,
  title={Perceptual compression for video storage and processing systems},
  author={Mazumdar, Amrita and Haynes, Brandon and Balazinska, Magda and Ceze, Luis and Cheung, Alvin and Oskin, Mark},
  booktitle={Proceedings of the ACM symposium on cloud computing},
  pages={179--192},
  year={2019}
}

@article{kaplanyan2019deepfovea,
  title={DeepFovea: Neural reconstruction for foveated rendering and video compression using learned statistics of natural videos},
  author={Kaplanyan, Anton S and Sochenov, Anton and Leimk{\"u}hler, Thomas and Okunev, Mikhail and Goodall, Todd and Rufo, Gizem},
  journal={ACM Transactions on Graphics (TOG)},
  volume={38},
  number={6},
  pages={1--13},
  year={2019},
  publisher={ACM New York, NY, USA}
}

@article{itti2004automatic,
  title={Automatic foveation for video compression using a neurobiological model of visual attention},
  author={Itti, Laurent},
  journal={IEEE transactions on image processing},
  volume={13},
  number={10},
  pages={1304--1318},
  year={2004},
  publisher={IEEE}
}

@article{assran2025vjepa2,
  title={V-jepa 2: Self-supervised video models enable understanding, prediction and planning},
  author={Assran, Mido and Bardes, Adrien and Fan, David and Garrido, Quentin and Howes, Russell and Muckley, Matthew and Rizvi, Ammar and Roberts, Claire and Sinha, Koustuv and Zholus, Artem and others},
  journal={arXiv preprint arXiv:2506.09985},
  year={2025}
}

@article{brown2020language,
  title={Language models are few-shot learners},
  author={Brown, Tom and Mann, Benjamin and Ryder, Nick and Subbiah, Melanie and Kaplan, Jared D and Dhariwal, Prafulla and Neelakantan, Arvind and Shyam, Pranav and Sastry, Girish and Askell, Amanda and others},
  journal={Advances in neural information processing systems},
  volume={33},
  pages={1877--1901},
  year={2020}
}

@inproceedings{li2025egom2p,
  title={Egom2p: Egocentric multimodal multitask pretraining},
  author={Li, Gen and Chen, Yutong and Wu, Yiqian and Zhao, Kaifeng and Pollefeys, Marc and Tang, Siyu},
  booktitle={Proceedings of the IEEE/CVF International Conference on Computer Vision},
  pages={10830--10843},
  year={2025}
}

@inproceedings{huang2018predicting,
  title={Predicting gaze in egocentric video by learning task-dependent attention transition},
  author={Huang, Yifei and Cai, Minjie and Li, Zhenqiang and Sato, Yoichi},
  booktitle={Proceedings of the European conference on computer vision (ECCV)},
  pages={754--769},
  year={2018}
}

@article{lai2022glc,
  title={In the eye of transformer: Global--local correlation for egocentric gaze estimation and beyond},
  author={Lai, Bolin and Liu, Miao and Ryan, Fiona and Rehg, James M},
  journal={International Journal of Computer Vision},
  volume={132},
  number={3},
  pages={854--871},
  year={2024},
  publisher={Springer}
}

@inproceedings{lai2024csts,
  title={Listen to look into the future: Audio-visual egocentric gaze anticipation},
  author={Lai, Bolin and Ryan, Fiona and Jia, Wenqi and Liu, Miao and Rehg, James M},
  booktitle={European Conference on Computer Vision},
  pages={192--210},
  year={2024},
  organization={Springer}
}

@inproceedings{kara2025diffeye,
  title={DiffEye: Diffusion-Based Continuous Eye-Tracking Data Generation Conditioned on Natural Images},
  author={Kara, Ozgur and Nisar, Harris and Rehg, James Matthew},
  booktitle={The Thirty-ninth Annual Conference on Neural Information Processing Systems}
}

@article{goodfellow2014generative,
  title={Generative adversarial nets},
  author={Goodfellow, Ian J and Pouget-Abadie, Jean and Mirza, Mehdi and Xu, Bing and Warde-Farley, David and Ozair, Sherjil and Courville, Aaron and Bengio, Yoshua},
  journal={Advances in neural information processing systems},
  volume={27},
  year={2014}
}

@article{kingma2014autoencoding,
  title={Auto-encoding variational bayes},
  author={Kingma, Diederik P and Welling, Max},
  journal={arXiv preprint arXiv:1312.6114},
  year={2013}
}

@inproceedings{lipman2023flow,
  title={Flow Matching for Generative Modeling},
  author={Lipman, Yaron and Chen, Ricky TQ and Ben-Hamu, Heli and Nickel, Maximilian and Le, Matthew},
  booktitle={The Eleventh International Conference on Learning Representations},
  year={2023}
}

@article{tong2024improving,
  title={Improving and generalizing flow-based generative models with minibatch optimal transport},
  author={Tong, Alexander and FATRAS, Kilian and Malkin, Nikolay and Huguet, Guillaume and Zhang, Yanlei and Rector-Brooks, Jarrid and Wolf, Guy and Bengio, Yoshua},
  journal={Transactions on Machine Learning Research}
}

@inproceedings{peebles2023dit,
  title={Scalable diffusion models with transformers},
  author={Peebles, William and Xie, Saining},
  booktitle={Proceedings of the IEEE/CVF international conference on computer vision},
  pages={4195--4205},
  year={2023}
}

@article{vaswani2017attention,
  title={Attention is all you need},
  author={Vaswani, Ashish and Shazeer, Noam and Parmar, Niki and Uszkoreit, Jakob and Jones, Llion and Gomez, Aidan N and Kaiser, {\L}ukasz and Polosukhin, Illia},
  journal={Advances in neural information processing systems},
  volume={30},
  year={2017}
}

@article{su2024roformer,
  title={Roformer: Enhanced transformer with rotary position embedding},
  author={Su, Jianlin and Ahmed, Murtadha and Lu, Yu and Pan, Shengfeng and Bo, Wen and Liu, Yunfeng},
  journal={Neurocomputing},
  volume={568},
  pages={127063},
  year={2024},
  publisher={Elsevier}
}

@article{hu2022lora,
  title={Lora: Low-rank adaptation of large language models},
  author={Hu, Edward J and Shen, Yelong and Wallis, Phillip and Allen-Zhu, Zeyuan and Li, Yuanzhi and Wang, Shean and Wang, Lu and Chen, Weizhu},
  journal={arXiv preprint arXiv:2106.09685},
  year={2021}
}

@article{palazzi2018predicting,
  title={Predicting the driver's focus of attention: The DR (eye) VE project},
  author={Palazzi, Andrea and Abati, Davide and Solera, Francesco and Cucchiara, Rita and others},
  journal={IEEE transactions on pattern analysis and machine intelligence},
  volume={41},
  number={7},
  pages={1720--1733},
  year={2018},
  publisher={IEEE}
}

@article{treisman1980feature,
  title={A feature-integration theory of attention},
  author={Treisman, Anne M and Gelade, Garry},
  journal={Cognitive psychology},
  volume={12},
  number={1},
  pages={97--136},
  year={1980},
  publisher={Elsevier}
}

@article{yarbus1967eye,
  title={Eye Movements and Vision},
  author={Yarbus, Alfred L},
  journal={Eye movements and vision.},
  pages={171},
  year={1967},
  publisher={Springer US}
}

@article{zhao2026reference,
  title={Reference Forcing: Long Video Generation with Region-Guided Drift Correction},
  author={Zhao, Sheng and Zheng, Haitian and Lin, Weikai and Zhao, Junjie and Zhu, Yuhao},
  journal={arXiv preprint arXiv:2611.21559},
  year={2026}
}

@article{zhao2026multidimensional,
  title={Multidimensional Observer Model and Perceptual Dimensions of Human Image Quality Assessment},
  author={Zhao, Sheng and Lin, Weikai and Zhu, Yuhao},
  journal={arXiv preprint arXiv:2609.38487},
  year={2026}
}

@article{land1999roles,
  title={The roles of vision and eye movements in the control of activities of daily living},
  author={Land, Michael and Mennie, Neil and Rusted, Jennifer},
  journal={Perception},
  volume={28},
  number={11},
  pages={1311--1328},
  year={1999},
  publisher={SAGE Publications Sage UK: London, England}
}

@article{rucci2018temporal,
  title={Temporal coding of visual space},
  author={Rucci, Michele and Ahissar, Ehud and Burr, David},
  journal={Trends in cognitive sciences},
  volume={22},
  number={10},
  pages={883--895},
  year={2018},
  publisher={Elsevier}
}

@article{rothkopf2007task,
  title={Task and context determine where you look},
  author={Rothkopf, Constantin A and Ballard, Dana H and Hayhoe, Mary M},
  journal={Journal of vision},
  volume={7},
  number={14},
  pages={16--16},
  year={2007},
  publisher={The Association for Research in Vision and Ophthalmology}
}

@article{henderson2017gaze,
  title={Gaze control as prediction},
  author={Henderson, John M},
  journal={Trends in cognitive sciences},
  volume={21},
  number={1},
  pages={15--23},
  year={2017},
  publisher={Elsevier}
}

@article{hayhoe2005eye,
  title={Eye movements in natural behavior},
  author={Hayhoe, Mary and Ballard, Dana},
  journal={Trends in cognitive sciences},
  volume={9},
  number={4},
  pages={188--194},
  year={2005},
  publisher={Elsevier}
}

@article{land2001roles,
  title={In what ways do eye movements contribute to everyday activities?},
  author={Land, Michael F and Hayhoe, Mary},
  journal={Vision research},
  volume={41},
  number={25-26},
  pages={3559--3565},
  year={2001},
  publisher={Elsevier}
}

@article{tatler2011eye,
  title={Eye guidance in natural vision: Reinterpreting salience},
  author={Tatler, Benjamin W and Hayhoe, Mary M and Land, Michael F and Ballard, Dana H},
  journal={Journal of vision},
  volume={11},
  number={5},
  pages={5--5},
  year={2011},
  publisher={The Association for Research in Vision and Ophthalmology}
}

@inproceedings{li2018eye,
  title={In the eye of beholder: Joint learning of gaze and actions in first person video},
  author={Li, Yin and Liu, Miao and Rehg, James M},
  booktitle={Proceedings of the European conference on computer vision (ECCV)},
  pages={619--635},
  year={2018}
}

@inproceedings{grauman2022ego4d,
  title={Ego4d: Around the world in 3,000 hours of egocentric video},
  author={Grauman, Kristen and Westbury, Andrew and Byrne, Eugene and Chavis, Zachary and Furnari, Antonino and Girdhar, Rohit and Hamburger, Jackson and Jiang, Hao and Liu, Miao and Liu, Xingyu and others},
  booktitle={Proceedings of the IEEE/CVF conference on computer vision and pattern recognition},
  pages={18995--19012},
  year={2022}
}

@article{bylinskii2019different,
  title={What do different evaluation metrics tell us about saliency models?},
  author={Bylinskii, Zoya and Judd, Tilke and Oliva, Aude and Torralba, Antonio and Durand, Fr{\'e}do},
  journal={IEEE transactions on pattern analysis and machine intelligence},
  volume={41},
  number={3},
  pages={740--757},
  year={2019},
  publisher={IEEE}
}

@article{chen2023analog,
  title={Analog bits: Generating discrete data using diffusion models with self-conditioning},
  author={Chen, Ting and Zhang, Ruixiang and Hinton, Geoffrey},
  journal={arXiv preprint arXiv:2208.04202},
  year={2022}
}

@article{itti1998saliency,
  title={A model of saliency-based visual attention for rapid scene analysis},
  author={Itti, Laurent and Koch, Christof and Niebur, Ernst},
  journal={IEEE Transactions on pattern analysis and machine intelligence},
  volume={20},
  number={11},
  pages={1254--1259},
  year={1998},
  publisher={Ieee}
}

@inproceedings{loshchilov2019decoupled,
  title     = {Decoupled Weight Decay Regularization},
  author    = {Loshchilov, Ilya and Hutter, Frank},
  booktitle = {International Conference on Learning Representations},
  year      = {2019},
}

@article{mennie2007lookahead,
  title={Look-ahead fixations: anticipatory eye movements in natural tasks},
  author={Neil Mennie and Mary M. Hayhoe and Brian T. Sullivan},
  journal={Experimental Brain Research},
  year={2007},
  volume={179},
  pages={427-442}
}

@article{ballard1995memory,
  title={Memory Representations in Natural Tasks},
  author={Dana H. Ballard and Mary M. Hayhoe and Jeff B. Pelz},
  journal={Journal of Cognitive Neuroscience},
  year={1995},
  volume={7},
  pages={66-80}
}

@article{kim2021gaze,
  title={Gaze-Based Dual Resolution Deep Imitation Learning for High-Precision Dexterous Robot Manipulation},
  author={Heecheol Kim and Yoshiyuki Ohmura and Yasuo Kuniyoshi},
  journal={IEEE Robotics and Automation Letters},
  year={2021},
  volume={6},
  pages={1630-1637}
}

@inproceedings{Judd_2012,
  title={A Benchmark of Computational Models of Saliency to Predict Human Fixations},
  author={Tilke Judd and Fr{\'e}do Durand and Antonio Torralba},
  year={2012}
}

@article{pellegrini2009you,
  title={You'll never walk alone: Modeling social behavior for multi-target tracking},
  author={Stefano Pellegrini and Andreas Ess and Konrad Schindler and Luc Van Gool},
  journal={2009 IEEE 12th International Conference on Computer Vision},
  year={2009},
  pages={261-268}
}

@article{ho2020denoising,
  title={Denoising Diffusion Probabilistic Models},
  author={Jonathan Ho and Ajay Jain and P. Abbeel},
  journal={ArXiv},
  year={2020},
  volume={abs/2006.11239}
}

@article{li2013learning,
  title={Learning to Predict Gaze in Egocentric Video},
  author={Yin Li and Alireza Fathi and James M. Rehg},
  journal={2013 IEEE International Conference on Computer Vision},
  year={2013},
  pages={3216-3223}
}

@article{Zhang_2017_CVPR,
  title={Deep Future Gaze: Gaze Anticipation on Egocentric Videos Using Adversarial Networks},
  author={Mengmi Zhang and Keng Teck Ma and Joo Hwee Lim and Qi Zhao and Jiashi Feng},
  journal={2017 IEEE Conference on Computer Vision and Pattern Recognition (CVPR)},
  year={2017},
  pages={3539-3548},
}

@article{itti2000saliency,
  title={A saliency-based search mechanism for overt and covert shifts of visual attention},
  author={Itti, Laurent and Koch, Christof},
  journal={Vision research},
  volume={40},
  number={10-12},
  pages={1489--1506},
  year={2000},
  publisher={Elsevier}
}

@inproceedings{zhao2025coordauth,
  title={Coordauth: Hands-free two-factor authentication in virtual reality leveraging head-eye coordination},
  author={Zhao, Sheng and Zhu, Junrui and Zhang, Shuning and Wang, Xueyang and Li, Hongyi and Yi, Fang and Yi, Xin and Li, Hewu},
  booktitle={2025 IEEE Conference Virtual Reality and 3D User Interfaces (VR)},
  pages={738--748},
  year={2025},
  organization={IEEE}
}

@article{lin2026lowpowar,
  title={LowPowAR: Power-Constrained Tone Mapping for Augmented Reality},
  author={Lin, Weikai and Zhao, Sheng and Ross, Ian and Marshall, Carl and Kondguli, Sushant and Zhu, Yuhao},
  journal={arXiv preprint arXiv:2607.19509},
  year={2026}
}

@inproceedings{liu2024grounding,
  title={Grounding dino: Marrying dino with grounded pre-training for open-set object detection},
  author={Liu, Shilong and Zeng, Zhaoyang and Ren, Tianhe and Li, Feng and Zhang, Hao and Yang, Jie and Jiang, Qing and Li, Chunyuan and Yang, Jianwei and Su, Hang and others},
  booktitle={European conference on computer vision},
  pages={38--55},
  year={2024},
  organization={Springer}
}

@article{ravi2024sam,
  title={SAM 2: Segment Anything in Images and Videos},
  author={Nikhila Ravi and Valentin Gabeur and Yuan-Ting Hu and Ronghang Hu and Chaitanya K. Ryali and Tengyu Ma and Haitham Khedr and Roman R{\"a}dle and Chlo{\'e} Rolland and Laura Gustafson and Eric Mintun and Junting Pan and Kalyan Vasudev Alwala and Nicolas Carion and Chao Wu and Ross B. Girshick and Piotr Doll'ar and Christoph Feichtenhofer},
  journal={ArXiv},
  year={2024},
  volume={abs/2408.00714}
}

@article{fuglede2004jensen,
  title={Jensen-Shannon divergence and Hilbert space embedding},
  author={Bent Fuglede and Flemming Tops{\o}e},
  journal={International Symposium onInformation Theory, 2004. ISIT 2004. Proceedings.},
  year={2004},
  pages={31},
}

@article{bai2025qwen3,
  title={Qwen3-vl technical report},
  author={Bai, Shuai and Cai, Yuxuan and Chen, Ruizhe and Chen, Keqin and Chen, Xionghui and Cheng, Zesen and Deng, Lianghao and Ding, Wei and Gao, Chang and Ge, Chunjiang and others},
  journal={arXiv preprint arXiv:2511.21631},
  year={2025}
}
\bibliographystyle{plainnat}

\newpage
\appendix
\section{Additional Visualization Results}
\label{app:visualizations}

This appendix gives the figures behind the analyses in \Sect{subsec: interpretation} and the failure cases in \Sect{subsec:limitation}.

\subsection{Per-Head Patterns of the Time-aware Self-Attention}
\label{app:head-attention}

\begin{figure}[!ht]
  \centering
  \includegraphics[width=\linewidth]{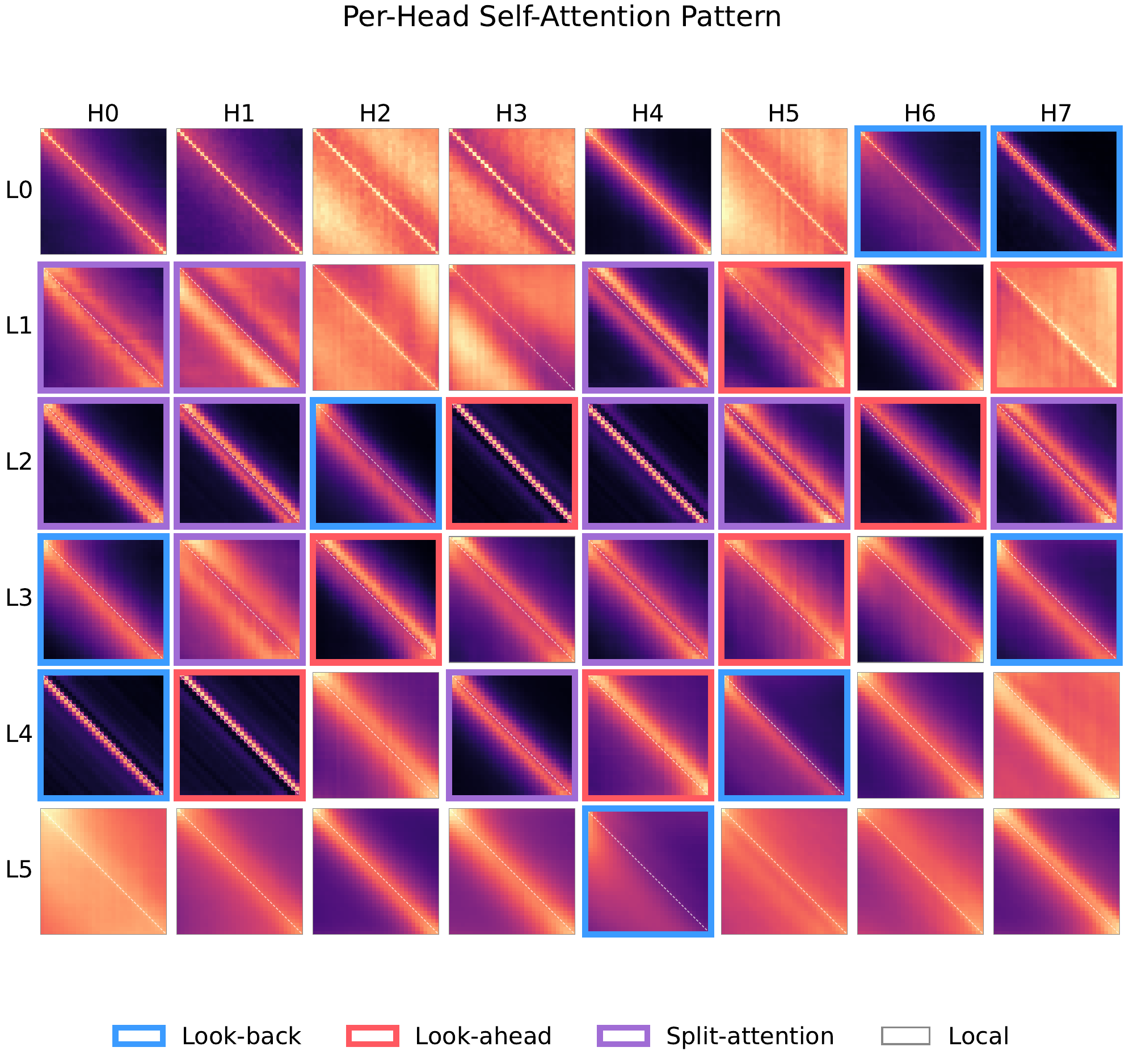}
  \caption{%
    \textbf{Per-head patterns of the time-aware self-attention aggregated across the EGTEA Gaze+ validation set.}
    Rows index denoiser layers, columns index attention heads.
    Each cell shows the canonical attention pattern (query frame on the vertical axis, source frame on the horizontal, both normalized to clip-relative position); brighter pixels correspond to higher attention weight after row-wise softmax.
    The dashed diagonal marks $\textit{source} = \textit{query}$, separating two halves with opposite temporal meaning: mass in the \emph{lower triangle} ($\textit{source} < \textit{query}$) is attention to \emph{past} frames (look-back), mass in the \emph{upper triangle} ($\textit{source} > \textit{query}$) is attention to \emph{future} frames (look-ahead), and mass on both sides marks split-attention heads with peaks straddling the current frame.
    Cell-frame color encodes this role: look-back (blue), look-ahead (red), split-attention (purple), local (grey).
  }
  \label{fig:self_attn_grid}
\end{figure}

\begin{figure}[!t]
  \centering
  \includegraphics[width=\linewidth]{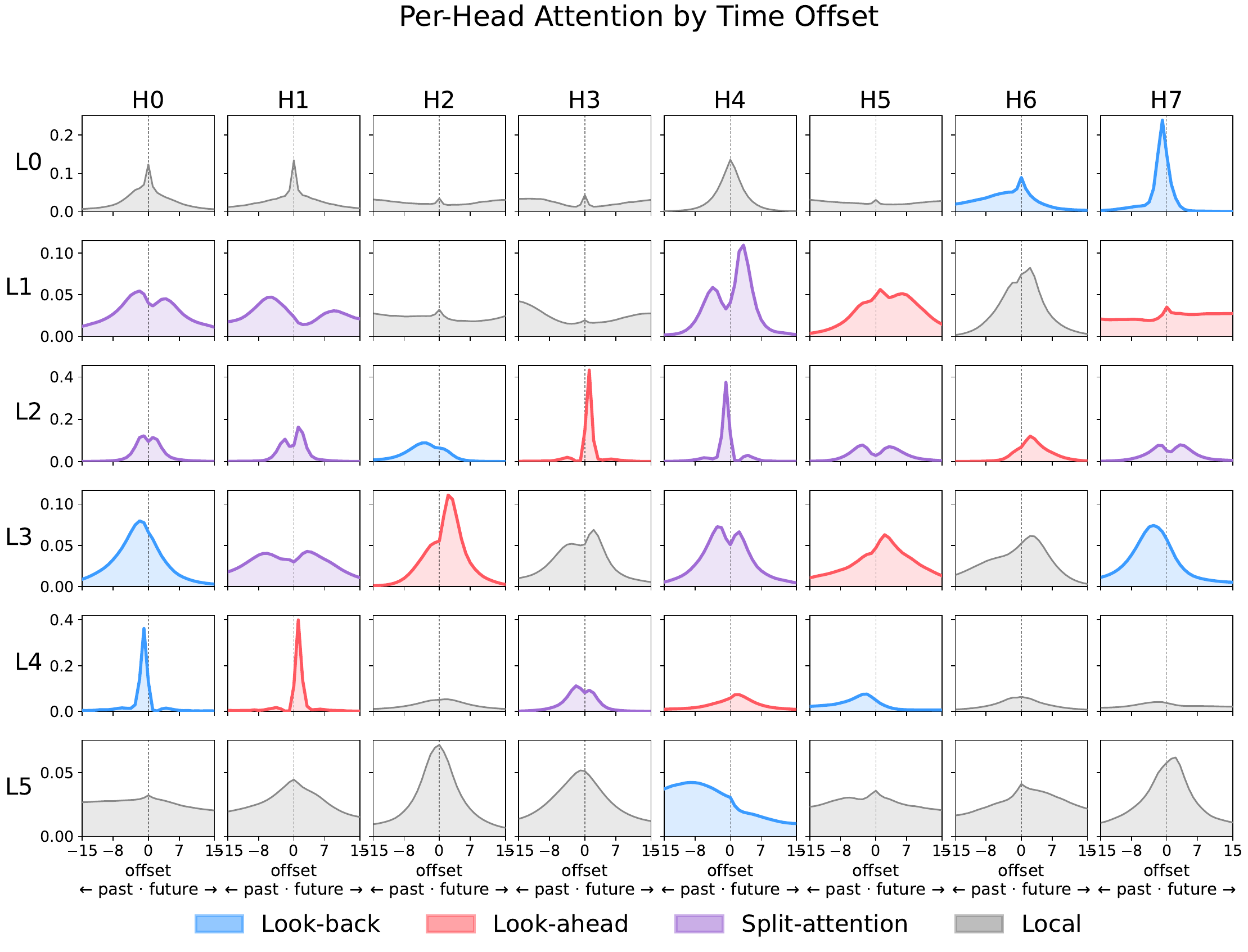}
  \caption{%
    \textbf{Per-head time-aware self-attention as a function of temporal offset.}
    Same heads as Fig.~\ref{fig:self_attn_grid}, summarized as a 1-D curve by averaging the attention pattern along each diagonal.
    The horizontal axis is the offset between source and query frame (negative = past, positive = future); the vertical dashed line marks zero offset.
    Look-ahead heads peak right of zero, look-back heads peak left, and split heads exhibit two peaks straddling zero with a clear valley at the current frame.
  }
  \label{fig:self_attn_offset}
\end{figure}

These two figures support the look-back and look-ahead heads described in \Sect{subsec: interpretation}.
We forward EGTEA Gaze+ validation clips through the trained main model at the midpoint of denoising and capture the gaze self-attention matrix of every (layer, head).
Because clip length varies, each matrix is resized to a common grid where both axes are normalized to clip-relative position; the per-head pattern shown is the average across clips.
We classify each head from its 1-D attention-vs-offset curve into one of four temporal roles: \emph{look-back}, \emph{look-ahead}, \emph{split-attention} (an inverted-M shape with peaks on both sides of the current frame), or \emph{local}; the same color scheme is used in both figures and the legend.
Figure~\ref{fig:self_attn_grid} shows the full 2-D pattern per head: look-back heads light up the lower triangle, look-ahead heads light up the upper triangle, and split heads light up both triangles.
Figure~\ref{fig:self_attn_offset} collapses the same data to a 1-D temporal-offset curve, where the three role types are immediately distinguishable from peak position.
\clearpage

\subsection{Per-Head Semantic Attribution}
\label{app:head-semantics}

\begin{figure}[!ht]
  \centering
  \includegraphics[width=0.78\linewidth]{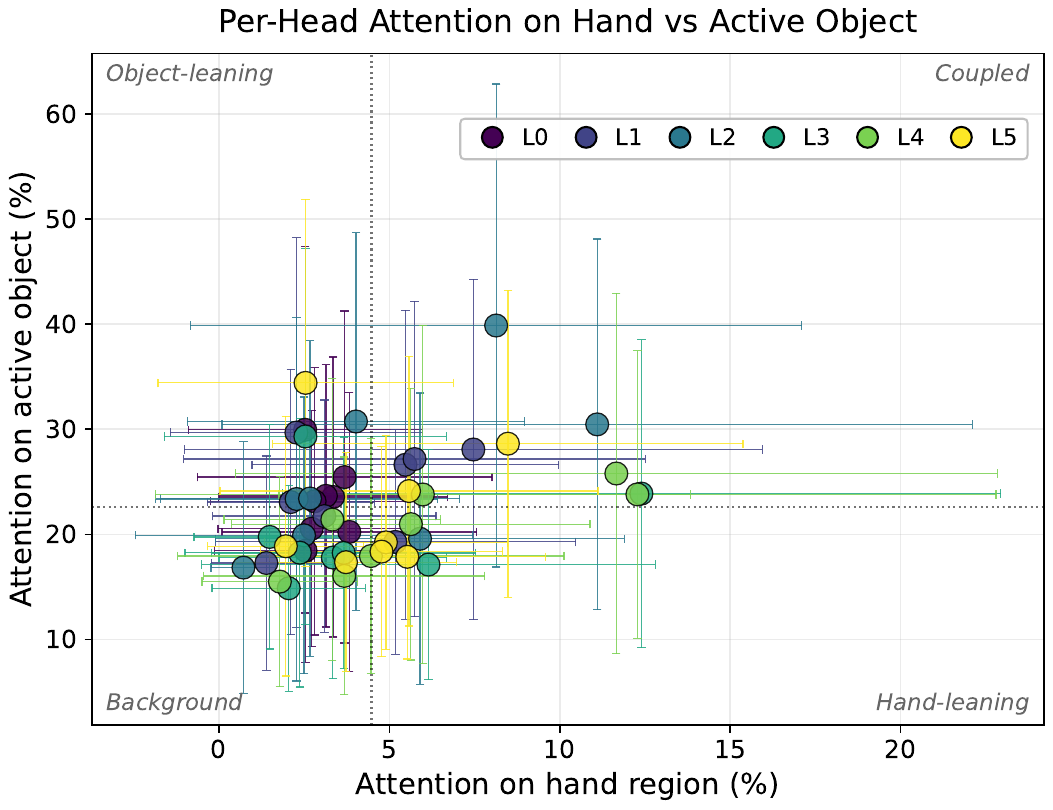}
  \caption{%
    \textbf{Hand vs.\ active-object attention mass per (layer, head)} on the EGTEA Gaze+ validation set.
    Each marker is one of the $48$ visual-cross-attention heads, plotted by mean attention mass on the hand (x-axis) and active-object (y-axis) regions; markers are colored by layer, error bars are clip-wise standard deviation.
    Dotted lines mark the global means (hand $=4.5\%$, object $=22.6\%$).
    Heads in the lower-right quadrant focus on the hand, those in the upper-left on the active object, and those in the upper-right couple both.
  }
  \label{fig:head_semantic_taxonomy}
\end{figure}

\begin{figure}[!ht]
  \centering
  \includegraphics[width=\linewidth]{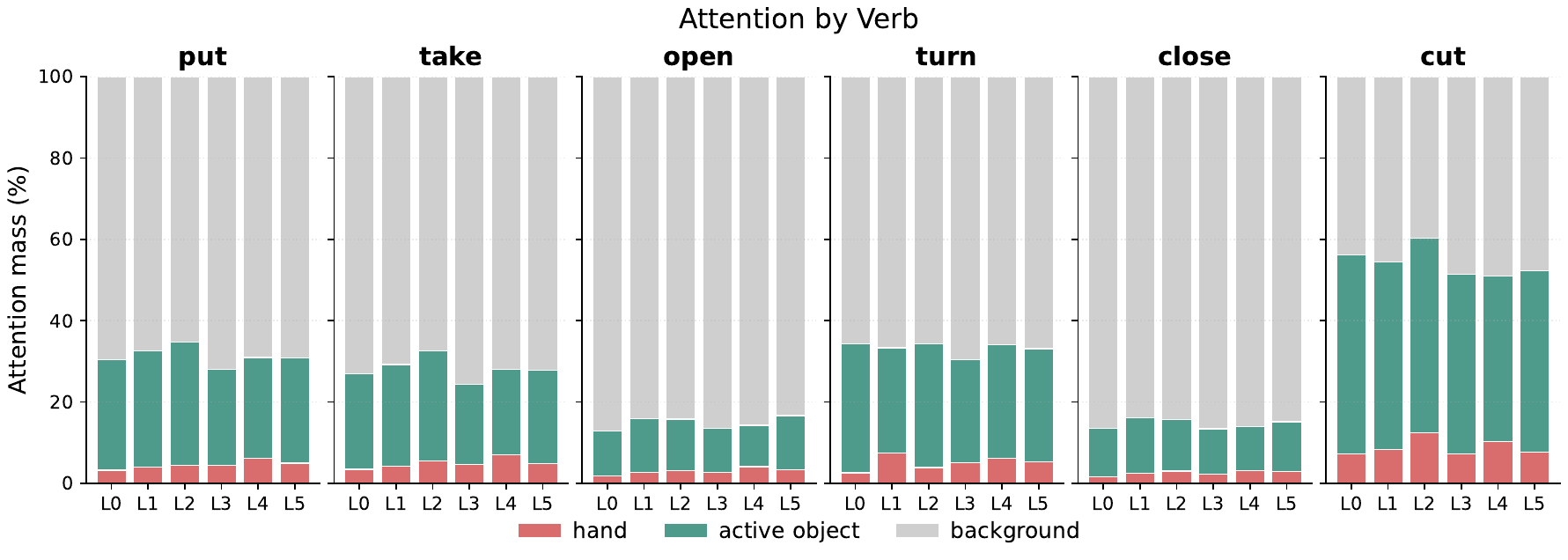}
  \caption{%
    \textbf{Layer-wise hand / active-object / background attention mass conditioned on verb.}
    For each of the six most frequent EGTEA verbs we average the per-region attention mass across heads at each layer.
    Manipulation verbs (``cut'', ``put'', ``take'', ``turn'') keep $25$--$50\%$ of the mass on the active object at every layer, while ``open'' and ``close'' route most mass to the background regardless of layer.
  }
  \label{fig:per_verb_breakdown}
\end{figure}

These two figures support the hand-focused and object-focused heads described in \Sect{subsec: interpretation}.
For each clip in the EGTEA Gaze+ validation set we extract per-frame Hand and Active-Object masks via GroundingDINO~\citep{liu2024grounding} and SAM~2~\citep{ravi2024sam}, downsampled to the $16{\times}16$ V-JEPA~2 token grid.
We then run a forward pass of the trained main model and capture the per-(layer, head) attention weights of the per-frame visual cross-attention sub-layer (\Sect{sec:method_cfm}) over the spatial tokens.
For each clip, layer, and head we sum the attention mass inside the hand mask, inside the active-object mask, and outside both (background); the three values are normalized to $1$ per (clip, layer, head) and then averaged across clips.
Figure~\ref{fig:head_semantic_taxonomy} plots each head as one point in the (hand, object) attention plane: heads in the lower-right quadrant concentrate on the hand, those in the upper-left on the active object, and those in the upper-right couple both, all relative to the global means (hand $=4.5\%$, object $=22.6\%$).
Figure~\ref{fig:per_verb_breakdown} groups clips by the first verb of their EGTEA action label, and the pattern differs across verbs.
``cut'', ``put'', ``take'', and ``turn'' act on an object already in hand, and the heads keep $25$--$50\%$ of the mass on the active object across all six layers.
``open'' and ``close'' act on a fixture such as a drawer, lid, or container, and the heads route most of the mass to the background regardless of layer.

\subsection{Failure Case Analysis}
\label{app:failure}

\begin{figure}[!ht]
  \centering
  \includegraphics[width=\linewidth]{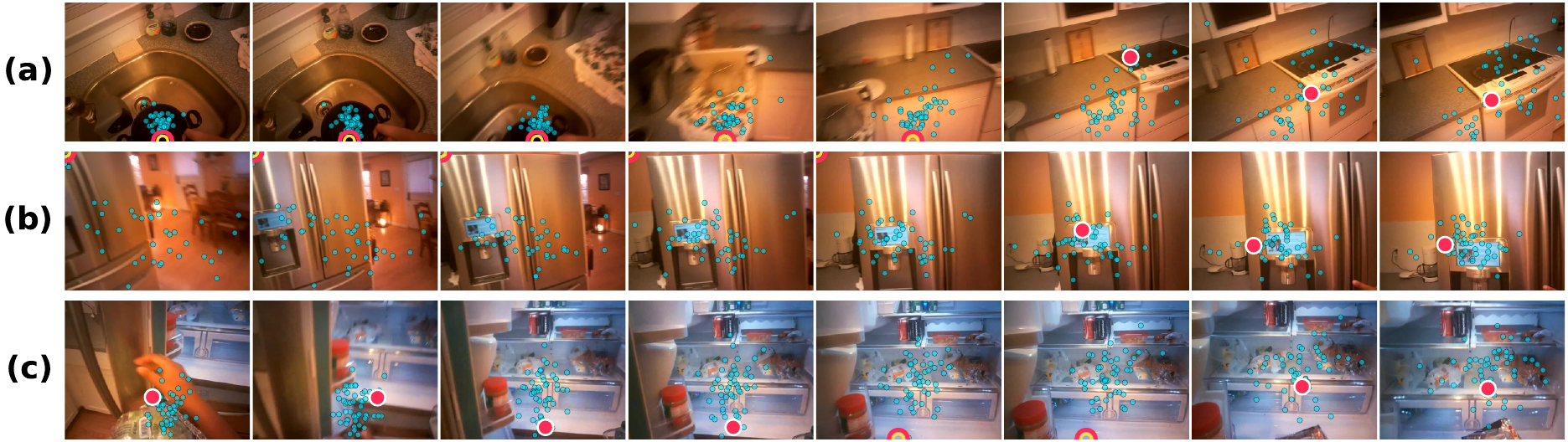}
  \caption{%
    \textbf{Failure cases.}
    Three rows from two validation clips, each row showing $8$ evenly-spaced frames along time.
    Cyan dots are the $K{=}50$ \proj{} predictions at that frame; the filled red circle is the recorded ground-truth gaze when it lies within the frame, and the hollow red ring with a yellow halo flags frames where the recording is clamped to the image boundary (off-frame).
    When the wearer saccades to a target outside the camera's view, the predicted cloud stays on the visible scene while the ground-truth marker drifts to the boundary.
  }
  \label{fig:failure_cases}
\end{figure}

\proj's failure cases concentrate on clips containing saccades whose endpoint lies outside the camera's field of view (\Fig{fig:failure_cases}).
A large saccade traverses tens of visual degrees in $30$--$80$\,ms~\citep{rucci2018temporal}, so at our $24$--$30$\,fps recording rate the eye can leave the captured frame within a single sample.
Once the eye target is outside the camera field of view (FOV), the model has no visual evidence on which to ground a prediction: \proj{} is conditioned solely on the recorded video, and the saccade endpoint by definition was never captured by it.
The predicted trajectories therefore stay on objects that remain visible, while the eye tracker reports a gaze position that is clamped to the image boundary.
Closing this gap would require a wider-FOV camera that captures the eventual saccade target, or a saccade-aware model that recognizes when the gaze has left the camera view, both of which we leave to future work.

\clearpage

\section{Conditional Flow Matching: Derivation}
\label{app:flow}

This appendix shows that the conditional flow matching loss $\mathcal{L}_{\mathrm{CFM}}$ used in \Sect{sec:training_inference} (Eq.~\ref{eq:method:cfm_loss}) is a sample-based $L^2$ regression on a velocity field whose global minimum yields, at the population level, $q_\theta = p^\star$.
\Sect{app:flow:objective} shows that minimizing $\mathcal{L}_{\mathrm{CFM}}$ recovers $p^\star$, and \Sect{app:perframe_gap} shows that per-frame factorization cannot recover the joint distribution.

\subsection{\texorpdfstring{Why $\mathcal{L}_{\mathrm{CFM}}$ recovers $p^\star$}{Why the CFM loss recovers the target distribution}}
\label{app:flow:objective}

For a noise sample $\mathbf{x}_0 \sim \mathcal{N}(\mathbf{0}, \mathbf{I}_{2T})$ and a trajectory $\mathbf{g} \sim p^\star(\cdot \mid \mathcal{V})$, the linear interpolant
\begin{equation}
  \mathbf{x}_s \;=\; (1 - s)\, \mathbf{x}_0 + s\, \mathbf{g}, \qquad s \in [0, 1],
  \label{eq:app:interpolant}
\end{equation}
starts at the Gaussian at $s = 0$ and ends at $p^\star$ at $s = 1$.
Its per-sample velocity is $\mathbf{g} - \mathbf{x}_0$.
Flow matching~\citep{lipman2023flow} shows that the marginal velocity field
\begin{equation}
  \mathbf{u}_s^\star(\mathbf{x} \mid \mathcal{V}) \;=\; \mathbb{E}\bigl[\mathbf{g} - \mathbf{x}_0 \,\big|\, \mathbf{x}_s = \mathbf{x},\, \mathcal{V}\bigr]
\end{equation}
moves the Gaussian to $p^\star$ along this path.
So if $\mathbf{v}_\theta = \mathbf{u}_s^\star$, integrating \Equ{eq:bg:cfm} from $\mathbf{x}_0$ gives $q_\theta = p^\star$.
The loss $\mathcal{L}_{\mathrm{CFM}}$ in \Eqn{eq:method:cfm_loss} regresses $\mathbf{v}_\theta$ onto $\mathbf{g} - \mathbf{x}_0$.
Because $\mathbf{v}_\theta$ sees only $(\mathbf{x}_s, s, \mathbf{c})$, the minimizer of this $L^2$ loss is the conditional expectation above, that is, $\mathbf{u}_s^\star$.
Minimizing $\mathcal{L}_{\mathrm{CFM}}$ therefore recovers $p^\star$ at the population level.
On EGTEA Gaze+, we multiply the loss by a per-frame validity mask $\mathbf{m}$ that removes blink frames; this does not change the minimizer on valid frames.

\subsection{Per-Frame Factorization Gap}
\label{app:perframe_gap}

Let $\mathcal{V}$ denote the video and $\mathbf{g}=(\mathbf{g}_1,\ldots,\mathbf{g}_T)$ the gaze trajectory, with ground-truth distribution $p^\star(\mathbf{g}\mid\mathcal{V})$.
A marginal model provides the trajectory distribution as
\begin{equation}
  q_{\mathrm{marg}}(\mathbf{g}\mid\mathcal{V}) \;=\; \prod_{t=1}^{T} q_t(\mathbf{g}_t\mid\mathcal{V}),
\end{equation}
where each $q_t$ is the parametric conditional distribution at frame $t$.
For any video $\mathcal{V}$, the best possible factorized approximation is obtained when each $q_t$ matches the corresponding true marginal $p^\star(\mathbf{g}_t\mid\mathcal{V})$.
Therefore,
\begin{equation}
\begin{aligned}
  \min_{q_1,\ldots,q_T} D_{\mathrm{KL}}\bigl(p^\star(\mathbf{g}\mid\mathcal{V}) \,\big\|\, q_{\mathrm{marg}}(\mathbf{g}\mid\mathcal{V})\bigr)
  &\;=\; D_{\mathrm{KL}}\Bigl(p^\star(\mathbf{g}\mid\mathcal{V}) \,\Big\|\, \prod_{t=1}^{T} p^\star(\mathbf{g}_t\mid\mathcal{V})\Bigr) \\
  &\;=\; \mathrm{TC}(\mathbf{g}_1,\ldots,\mathbf{g}_T\mid\mathcal{V}),
\end{aligned}
\end{equation}
where $\mathrm{TC}$ denotes total correlation, defined as the KL divergence between a joint distribution and the product of its marginals.
Since KL divergence is nonnegative, $\mathrm{TC}(\mathbf{g}_1,\ldots,\mathbf{g}_T\mid\mathcal{V})=0$ if and only if
\begin{equation}
  p^\star(\mathbf{g}\mid\mathcal{V}) \;=\; \prod_{t=1}^{T} p^\star(\mathbf{g}_t\mid\mathcal{V}),
\end{equation}
meaning that gaze positions are conditionally independent across time given the video.
Human gaze clearly contains temporal dependencies.
Therefore, even if a marginal model perfectly estimates every per-frame distribution, it can still have an irreducible error in the joint trajectory distribution.

\section{Velocity Network Architecture}
\label{app:architecture}

This section gives the parts of the velocity network $\mathbf{v}_\theta$ that \Sect{sec:method_cfm} does not cover: the encoder and projection (\Sect{app:vlm}), the rotary position embedding (\Sect{app:rope}), and the four sub-layers of a denoiser block (\Sect{app:block}).

\subsection{Encoder and Projection}
\label{app:vlm}

\paragraph{Backbone selection.}
We use V-JEPA~2 ViT-L because its features capture both fine-grained motion and clip-level action structure~\citep{assran2025vjepa2}: the former supplies the local visual evidence read by our per-frame cross-attention (bottom-up), and the latter the task-level structure pooled by our task-token bank (top-down).
A linear probe on frozen features supports this choice (Table~\ref{tab:probe_backbone}): at the native resolution $256{\times}256$, V-JEPA~2 matches Qwen3-VL-8B~\citep{bai2025qwen3} in AUC and exceeds the best Qwen3-VL layer in F1 (0.378 vs.\ 0.370) and Precision (0.292 vs.\ 0.280), while a scale-matched input $240{\times}320$ degrades every metric.

\begin{table}[h]
  \caption{%
    \textbf{Linear probing of frozen backbone features on EGTEA Gaze+}.
    Top block: V-JEPA~2 ViT-L at its native pre-training resolution and at a 4:3 input that matches the dataset aspect ratio.
    Bottom block: Qwen3-VL-8B at four LM layers, native resolution.
    All probes are ridge regressions on the frozen spatial token grid; per-column best in bold.}
  \label{tab:probe_backbone}
  \centering\small
  \begin{tabular}{l c c ccccc}
    \toprule
    Backbone & Resolution & Layer & AUC$\uparrow$ & F1$\uparrow$ & P$\uparrow$ & R$\uparrow$ & AAE$\downarrow$ \\
    \midrule
    \textbf{V-JEPA~2 ViT-L} & $\mathbf{256{\times}256}$ (native) & last & \textbf{0.944} & \textbf{0.378} & \textbf{0.292} & 0.535 & 11.39 \\
    V-JEPA~2 ViT-L          & $240{\times}320$ (matched)         & last & 0.913          & 0.296          & 0.216          & 0.469 & 12.87 \\
    \midrule
    Qwen3-VL-8B & native & 2  & 0.944 & 0.370 & 0.280 & \textbf{0.549} & 11.37 \\
    Qwen3-VL-8B & native & 4  & 0.944 & 0.370 & 0.280 & 0.547          & 11.37 \\
    Qwen3-VL-8B & native & 8  & 0.943          & 0.369 & 0.278 & \textbf{0.549} & 11.34 \\
    Qwen3-VL-8B & native & 16 & 0.942          & 0.368 & 0.279 & 0.541          & \textbf{11.24} \\
    \bottomrule
  \end{tabular}
\end{table}

\paragraph{Projection and spatial encoder.}
A linear layer projects each encoder token from $D{=}1024$ to $d{=}256$.
We add a 2D sinusoidal spatial position encoding and pass the tokens of each frame through a two-layer Transformer encoder, which lets the $N$ spatial tokens of a frame exchange information.
This encoder works on each frame on its own and never mixes information across time.

\subsection{Rotary Position Embedding}
\label{app:rope}

For the frame-to-frame self-attention (Sub-layer~1 below) we use 1D RoPE~\citep{su2024roformer} indexed by the frame position $p \in \{0, \ldots, T{-}1\}$, with base $\theta = 10000$.
RoPE rotates each query/key pair $(q_{2j}, q_{2j+1})$ by $\phi_{p,j} = p \cdot \theta^{-2j/d_h}$ before the attention dot-product; values are not rotated.
By construction the logit $a_{p,p'} = \langle \text{RoPE}(\mathbf{q}_p), \text{RoPE}(\mathbf{k}_{p'})\rangle / \sqrt{d_h}$ depends only on the relative offset $p - p'$, which is the structure gaze dynamics require (consecutive-frame coupling and saccade step length scale with $|p - p'|$, not with absolute $p$).
See \Sect{sec:ablation} and \Sect{sec:design} for the effect of RoPE.

\subsection{DenoiserBlock: Four Sub-layers}
\label{app:block}

Each denoiser block combines the tensors defined in the previous subsections: gaze tokens $\mathbf{h}$ from the input pipeline, projected encoder tokens $\mathbf{V}$ from Section~\ref{app:vlm}, and task tokens $\mathbf{c}_{\text{task}}$ from \Eqn{eq:method:task_bank}.
Each of the $L{=}6$ blocks applies four pre-norm residual sub-operations to the gaze tokens $\mathbf{h} \in \R^{T \times d}$.
All attentions use $H = 8$ heads with $d_h = d/H = 32$ and layer-specific QKV/output projections (multi-head attention, MHA, follows the standard definition).

\paragraph{Sub-layer 1: Frame-to-frame self-attention with RoPE.}
\begin{equation}
  \mathbf{h} \;\leftarrow\; \mathbf{h} + \text{MHA}\bigl(\text{RoPE}(\mathbf{Q}),\, \text{RoPE}(\mathbf{K}),\, \mathbf{V};\ \mathbf{h}\bigr),
  \label{eq:sa_out}
\end{equation}
where $\mathbf{Q}, \mathbf{K}, \mathbf{V}$ are linear projections of $\text{LayerNorm}(\mathbf{h})$ and RoPE is applied to queries and keys but not to values.
This is the only pathway along which gaze tokens at different frames exchange information, and is therefore where the joint structure of $p(\mathbf{g} \mid \mathcal{V})$ enters the network.

\paragraph{Sub-layer 2: Per-frame visual cross-attention.}
Each gaze token $\mathbf{h}_t$ cross-attends \emph{only} to its own frame's $N$ spatial tokens $\mathbf{V}_t$:
\begin{equation}
  \mathbf{h}_t \;\leftarrow\; \mathbf{h}_t + \text{MHA}\bigl(\text{LayerNorm}(\mathbf{h}_t),\, \mathbf{V}_t,\, \mathbf{V}_t\bigr).
  \label{eq:ca_out}
\end{equation}
Long-range temporal aggregation is left to Sub-layer~1; this sub-layer reads the encoder tokens of the frame being denoised.
Unlike Sub-layer~3, this sub-layer has no gate, so the visual condition acts at full strength from the start of training.

\paragraph{Sub-layer 3: Task cross-attention.}
\label{app:task-cross-atten}
Each gaze token cross-attends to the clip-level task tokens $\mathbf{c}_{\text{task}} \in \R^{(M+1) \times d}$ defined in \Eqn{eq:method:task_bank}:
\begin{equation}
  \mathbf{h} \;\leftarrow\; \mathbf{h} + \sigma(g^{(\ell)})\cdot \text{MHA}\bigl(\text{LayerNorm}(\mathbf{h}),\, \mathbf{c}_{\text{task}},\, \mathbf{c}_{\text{task}}\bigr).
  \label{eq:task_xattn}
\end{equation}
The per-layer gate $g^{(\ell)} \in \R^d$ is a channel-wise learnable vector applied element-wise to the cross-attention residual (broadcast over the $T$ gaze tokens).
All channels are initialized at $-2$ so that $\sigma(g^{(\ell)}_i) \approx 0.12$ uniformly at the start of training: the task input starts weak and grows as the Task Injection module learns useful features, letting the rest of the network train without interference from a randomly initialized task pathway.

\section{Training and Sampling Algorithms}
\label{app:tricks}

This section gives the training and inference algorithm (\Apx{app:alg}), self-conditioning (\Apx{app:selfcond}), and the sliding window for long clips (\Apx{app:window}).

\subsection{Training and Inference Algorithm}
\label{app:alg}

Algorithm~\ref{alg:gazeflow} summarizes training and inference.
For clarity, it omits self-conditioning (\Apx{app:selfcond}) and the parameter EMA.

\begin{algorithm}[t]
\caption{Training and inference for \ours{}.}
\label{alg:gazeflow}
\begin{algorithmic}[1]
\Require Dataset $\mathcal{D}$; encoder $E_\psi$; velocity network $\mathbf{v}_\theta$; Euler steps $S$; samples per clip $K$.
\Statex \textbf{Training.}
\Repeat
  \State Sample $(\mathcal{V}, \mathbf{g}, \mathbf{m}) \sim \mathcal{D}$; \; $\mathbf{c} \gets E_\psi(\mathcal{V})$.
  \State Sample $s \sim U[0,1]$ and $\mathbf{x}_0 \sim \mathcal{N}(\mathbf{0}, \mathbf{I})$; \; $\mathbf{x}_s \gets (1-s)\,\mathbf{x}_0 + s\,\mathbf{g}$.
  \State $\mathcal{L} \gets \| \mathbf{m} \odot (\mathbf{v}_\theta(\mathbf{x}_s, s, \mathbf{c}) - (\mathbf{g} - \mathbf{x}_0)) \|_2^2$; \; update $(\psi, \theta)$.
\Until{convergence}
\Statex \textbf{Inference.}
\State $\mathbf{c} \gets E_\psi(\mathcal{V})$. \Comment{once per clip}
\For{$k = 1, \ldots, K$ \textbf{in parallel}}
  \State $\mathbf{x}^{(k)} \sim \mathcal{N}(\mathbf{0}, \mathbf{I})$.
  \For{$i = 0, \ldots, S-1$}
    \State $\mathbf{x}^{(k)} \gets \mathbf{x}^{(k)} + \frac{1}{S}\, \mathbf{v}_\theta(\mathbf{x}^{(k)}, i/S, \mathbf{c})$.
  \EndFor
\EndFor
\State \Return $\{\mathbf{x}^{(k)}\}_{k=1}^{K}$.
\end{algorithmic}
\end{algorithm}

\subsection{Self-Conditioning Details}
\label{app:selfcond}

Self-conditioning~\citep{chen2023analog} feeds the model's previous estimate of the clean trajectory, $\hat{\mathbf{x}}_1 = \mathbf{x}_s + (1 - s)\, \mathbf{v}_\theta$, back as an extra input.
During training, we use this estimate on half of the steps and a zero input on the other half.
At inference, each Euler step uses the estimate from the previous step, starting from zero.

\subsection{Sliding-Window Inference for Long Clips}
\label{app:window}

The velocity network is trained on fixed windows of $W = 64$ frames.
At inference, clips longer than $W$ are split into overlapping windows of stride $W_s = W/2 = 32$, and Algorithm~\ref{alg:gazeflow} is run independently inside each window.
In the overlap region between consecutive windows $i$ and $i{+}1$, predictions are linearly blended along the frame axis $\tau$ to avoid window seams:
\begin{equation}
  \mathbf{x}_\tau^{\text{blend}}
    \;=\; (1 - \alpha_\tau)\, \mathbf{x}_\tau^{(i)} + \alpha_\tau\, \mathbf{x}_\tau^{(i+1)},
  \qquad
  \alpha_\tau \;=\; \frac{\tau - \tau_{\text{start}}^{(i+1)}}{W_s} \in [0, 1],
\end{equation}
where $\tau_{\text{start}}^{(i+1)}$ is the first frame of window $i{+}1$ and $\alpha_\tau$ ramps from $0$ to $1$ across the overlap.
We pad clips shorter than $W$ by repeating the last frame, and we cut the output back to the original length.

\section{Experimental Details}
\label{app:implementation}

This appendix supplements \Sect{sec:setup}: \Sect{app:setup_datasets} describes the datasets, \Sect{app:setup_baselines} the baselines, \Sect{app:setup_metrics} the metrics and evaluation protocol, and \Sect{app:setup_training} our hyperparameters.

\subsection{Datasets}
\label{app:setup_datasets}

\paragraph{EGTEA Gaze+.}
EGTEA Gaze+~\citep{li2018eye} contains 86 cooking videos (106 action classes with 19 verbs and 53 nouns) recorded at $640\times480$ / 24\,fps together with synchronized gaze from an SMI head-mounted eye tracker at 30\,Hz.
We follow the official train / val split from~\citet{li2018eye}, yielding 8{,}299 training and 2{,}022 validation clips after aligning video and gaze timelines.
Eye-tracker samples are classified into Fixation, Saccade, and Blink events by the authors' BeGaze output; we carry forward all event types into the trajectory target but exclude Blink frames from the loss via a binary validity mask $\mathbf{m}$.

\paragraph{Ego4D (Aria-glass gaze subset).}
The Ego4D~\citep{grauman2022ego4d} benchmarks include a subset of clips captured with Project Aria glasses equipped with an integrated eye tracker, providing per-frame gaze at $1088\times1080$ / 30\,fps.
We use the released train / val partition (12{,}178 / 5{,}202 clips).
Raw gaze is already aligned to the video grid by the Aria SDK; 
we rescale coordinates to the native $1088\times1080$ resolution, normalize to $[-1,1]$ for training.
Ego4D has no action or verb labels, so the task-token probe in \Sect{sec:task_probe} uses EGTEA Gaze+ only.

\paragraph{Windowing.}
All experiments operate on fixed $T=64$-frame windows drawn from each clip: clips shorter than $T$ are right-padded with the last valid frame and their padding positions are zeroed in the validity mask; clips longer than $T$ are trained with random crops and evaluated with a sliding window (stride $32$, linear blend in the overlap, Appendix~\ref{app:window}).
Clip-length statistics and per-split window counts are summarized in Table~\ref{tab:dataset_stats}.

\begin{table}[h]
  \centering
  \scriptsize
  \caption{Dataset summary.}
  \label{tab:dataset_stats}
  \begin{tabular}{@{}l ccccc@{}}
    \toprule
    Dataset & Resolution & fps & Train clips & Val clips \\
    \midrule
    EGTEA Gaze+       & $640\times480$     & 24 & 8{,}299  & 2{,}022 \\
    Ego4D (Aria gaze) & $1088\times1080$   & 30 & 12{,}178 & 5{,}202  \\
    \bottomrule
  \end{tabular}
\end{table}

\subsection{Baselines}
\label{app:setup_baselines}

\paragraph{Attention Transition (AT).}
AT~\citep{huang2018predicting} is trained in three stages: (SP) a two-stream CNN predicts a per-frame heatmap, (LSTM) a recurrent module classifies each frame as Fixation or Saccade, and (LF) a late-fusion CNN re-weights SP by the fixation--saccade posterior.
On EGTEA we retrain the full SP$+$LSTM$+$LF pipeline with the authors' released recipe on the 8{,}299 train clips.
On Ego4D, fixation/saccade event labels are not provided, so only Stage~SP is retrained and the LSTM and LF stages are disabled.
AT inference heatmaps are resized to $480\times640$ before evaluation.

\paragraph{GLC.}
GLC~\citep{lai2022glc} is a transformer saliency model with a ``global--local correlation'' head; we use the authors' released checkpoints for EGTEA and Ego4D without modification.
Predictions are produced at $64\times64$ and upscaled to $480\times640$ for evaluation to match our saliency protocol.

\paragraph{EgoM2P.}
EgoM2P~\citep{li2025egom2p} is an autoregressive multimodal foundation model pretrained on eight egocentric datasets that emits discrete gaze tokens one at a time.
We report two variants: (i)~the released zero-shot checkpoint, and (ii)~a LoRA-finetuned variant with rank $r=16$ applied to the query and value projections of all transformer layers, trained on each dataset's train split with the same optimizer as our own model.
For both variants we draw $K{=}50$ trajectories per clip at temperature $1.0$, matching our sampler.

\paragraph{Prior-only references.}
Two simple priors serve as reference points in \Tbl{tab:main} and \Tbl{tab:dynamics}.
\emph{Center Bias} is a static 2D Gaussian centered at the image center with per-axis standard deviation matched to the empirical fixation spread of each dataset's train split; it is evaluated with the same saliency protocol as every other baseline and has no trainable parameters.
\emph{Random Walk} takes the previous frame's ground-truth fixation (or the clip centroid at $t{=}0$) and samples the next predicted fixation from an isotropic Gaussian with $\sigma$ matched to the empirical frame-to-frame displacement of the train split; it is explicitly a first-order Markov prior and carries no visual input at all.

\subsection{Metrics and Evaluation Protocol}
\label{app:setup_metrics}

The metrics below mirror the two-axis structure of \Sect{sec:setup}: \emph{per-frame prediction accuracy} scores each predicted heatmap against the per-frame ground-truth gaze, while \emph{alignment with human gaze temporal dynamics} scores sampled trajectories against the human trajectory either directly or as a step-size distribution.

\paragraph{Per-frame prediction accuracy.}
In egocentric gaze datasets, each frame is annotated with a ground-truth gaze point.
Following the GLC protocol~\citep{lai2022glc}, we place a Gaussian kernel at this point to form a fixation map, which is then compared with the predicted heatmap.
\emph{Area Under the Curve} (AUC, the AUC-Judd variant~\citep{bylinskii2019different, Judd_2012}) treats the predicted heatmap as a continuous per-pixel score and measures how well it ranks ground-truth gaze locations above non-gaze locations.
\emph{F1}, \emph{Precision}, and \emph{Recall} instead binarize the predicted heatmap and the ground-truth fixation map at a fixed threshold and measure the overlap between the two.
\emph{Average Angular Error} (AAE)~\citep{huang2018predicting} measures the angular deviation between the predicted gaze point and the ground-truth gaze point.

Per-frame metrics aggregate hierarchically: per frame, then averaged within each clip (the clip-mean), then averaged across clips (the global mean), which follows GLC's original protocol and makes each clip count equally regardless of length.
For F1, Precision, and Recall, the threshold is selected by sweeping over the heatmap; following~\citet{lai2022glc}, we report Adaptive F1 (the best F1 over the sweep) with Precision and Recall taken at the same threshold.
Beyond AUC-Judd, we additionally report four standard saliency metrics following~\citet{bylinskii2019different}: NSS (the mean normalized saliency at fixation locations), the linear correlation coefficient CC, the histogram similarity SIM, and the KL divergence from the ground-truth to the predicted distribution.

\paragraph{Alignment with human gaze temporal dynamics.}
Per-frame metrics score each frame independently and therefore do not measure whether the predicted gaze forms a natural scanpath over time.
The metrics below evaluate how closely the dynamics of sampled trajectories match those of human gaze.
\emph{Average Displacement Error} (ADE) and \emph{Dynamic Time Warping distance} (DTW)~\citep{pellegrini2009you, kara2025diffeye} compare each predicted trajectory against the human trajectory directly; DTW is the minimum Euclidean alignment cost between the two under a monotone temporal warping, length-normalised by the trajectory length.
For both, we report the mean over $K=50$ samples and the best-of-$K$ value.

We additionally compare step-size statistics.
For each trajectory, we measure how far the gaze moves between consecutive frames, and build a displacement histogram, which represents the empirical distribution of these per-frame distances, aggregated across all frames and clips.
We report the mean and median displacement as ratios to the human reference, and the Jensen--Shannon divergence (JSD) between the predicted and human displacement histograms.
The JSD uses base $e$ on a $100$-bin histogram over the union range of all method and human displacements, which captures distribution shape rather than only its first two moments.

\paragraph{Trajectory-to-heatmap conversion.}
\label{app:trajectory2sal}
Trajectory generators -- our model, EgoM2P, and pseudo-trajectory baselines -- are converted to per-frame heatmaps so that all methods are comparable on the saliency metrics above.
For each frame $t$ we (i)~denormalize all $K$ predicted gaze points from $[-1,1]$ to pixel coordinates (the native evaluation resolution of the target dataset); (ii)~place a Gaussian kernel with $\sigma=25$\,px at each prediction; (iii)~average over the $K$ samples to obtain $\mathbf{S}_t$.
The bandwidth $\sigma{=}25$\,px matches the one used by the released GLC predictions.

\paragraph{Frozen-encoder protocol for ablations.}
The ablation study (Section~\ref{sec:ablation}) and the design-choice analysis (Section~\ref{sec:design}) evaluate the frozen-encoder variant of \proj{} on a fixed validation subset, rather than the LoRA variant on the full validation split.
We make this choice for two reasons.
First, end-to-end LoRA training together with $K{=}50$ sampling for every architectural variant is computationally prohibitive.
Second, every component varied in those studies (the visual and task cross-attention pathways, the task-token bank, the temporal RoPE, and the flow-matching head) operates on top of the encoder representation and does not change the encoder itself.
We therefore expect the ranking of variants to hold for the LoRA-adapted encoder as well.

\paragraph{Validation subset.}
The subset is a random sampled subset of the EGTEA Gaze+ validation split, drawn once with a fixed random seed and held identical across every variant in the ablation and design-choice studies.
After applying the GLC-format Fixation filter, the subset contains 
$1{,}536$ frame-level paired observations per metric.

\paragraph{Significance testing.}
We compare each result in \Tbl{tab:main}, \Tbl{tab:ablation}, and \Sect{sec:design} with its reference using a paired Wilcoxon signed-rank test on the frame-level scores, since these scores are heavy-tailed.
Almost every comparison is significant at $p < 0.05$; the exception is \emph{Recall} on Ego4D in \Tbl{tab:main}, where Random Walk is not beaten by \proj{}.

\subsection{Training and Sampling Hyperparameters}
\label{app:setup_training}

\paragraph{Video feature extraction.}
The visual encoder is V-JEPA~2 ViT-L~\citep{assran2025vjepa2}, fine-tuned with LoRA jointly with the velocity network (``LoRA configuration'' below).
Frames are resized to $256\times256$ and read at $24$\,fps on EGTEA Gaze+ and $30$\,fps on Ego4D, and we take the last hidden state of the encoder.
This yields $N=256$ spatial tokens of dimension $D=1024$ per frame.

\paragraph{Optimizer and schedule.}
AdamW with $\beta_1{=}0.9$, $\beta_2{=}0.999$, weight decay $10^{-4}$; learning rate $10^{-4}$ with cosine annealing to $10^{-6}$ over 80 epochs; effective batch size $16$; EMA decay $\beta_{\text{EMA}}{=}0.995$, and all evaluations use the EMA weights; gradient clipping at norm $1.0$.
Early stopping uses patience $15$ on validation loss computed at end of each epoch.

\paragraph{LoRA configuration.}
The V-JEPA~2 ViT-L encoder is fine-tuned with LoRA~\citep{hu2022lora} rank $r{=}16$, $\alpha{=}32$, dropout $0.05$, applied to the query and value projections of every transformer block.
All other encoder weights are frozen; only LoRA adapters, the input projection $\mathbf{W}_{\text{proj}}$, and the velocity network are trainable.

\paragraph{Flow-matching parameters.}
We use the OT-form conditional interpolant $\mathbf{x}_s = (1 - (1-\sigma_{\min}) s)\,\mathbf{x}_0 + s\,\mathbf{g}$ with $\sigma_{\min}=10^{-3}$, which differs from the linear interpolant in Eq.~\eqref{eq:app:interpolant} only by an $O(\sigma_{\min})$ correction below pixel precision; self-conditioning is applied with probability $0.5$ during training and deterministically on at inference; training uses the uniform schedule $s\sim\mathcal{U}(0,1)$ as in~\citet{lipman2023flow}.

\paragraph{Sampling.}
Euler ODE integration with $S=50$ steps and $K=50$ trajectories per clip.
Sliding window $W=64$, stride $32$, linear blend in overlap (Appendix~\ref{app:window}).

\section{Additional Results}
\label{app:additional_results}

This appendix gives results that support \Sect{sec:experiments}: the full main comparison (\Sect{app:full_main_table}), the inference cost and the effect of the sample count $K$ and step count $S$ (\Sect{app:efficiency}), the task-query sweep (\Sect{app:k_sweep}), and two further ablations (\Sect{app:ablations_extra}).

\subsection{Full Main Comparison Table}
\label{app:full_main_table}

Table~\ref{tab:main_full} expands Table~\ref{tab:main} with all nine standard metrics, split into the saliency block (NSS, AUC, CC, KL, SIM) and the localization block (F1, P, R, AAE).

\begin{table}[h]
  \caption{Full saliency and localization comparison on EGTEA Gaze+ and Ego4D.}
  \label{tab:main_full}
  \centering
  \scriptsize
  \setlength{\tabcolsep}{1.6pt}
  \resizebox{\linewidth}{!}{%
  \begin{tabular}{@{}l ccccc cccc ccccc cccc@{}}
    \toprule
    & \multicolumn{9}{c}{\textbf{EGTEA Gaze+}} & \multicolumn{9}{c}{\textbf{Ego4D}} \\
    \cmidrule(lr){2-10} \cmidrule(lr){11-19}
    & \multicolumn{5}{c}{Saliency} & \multicolumn{4}{c}{Localization}
    & \multicolumn{5}{c}{Saliency} & \multicolumn{4}{c}{Localization} \\
    \cmidrule(lr){2-6} \cmidrule(lr){7-10} \cmidrule(lr){11-15} \cmidrule(lr){16-19}
    Method
      & NSS$\uparrow$ & AUC$\uparrow$ & CC$\uparrow$ & KL$\downarrow$ & SIM$\uparrow$
      & F1$\uparrow$ & P$\uparrow$ & R$\uparrow$ & AAE$\downarrow$
      & NSS$\uparrow$ & AUC$\uparrow$ & CC$\uparrow$ & KL$\downarrow$ & SIM$\uparrow$
      & F1$\uparrow$ & P$\uparrow$ & R$\uparrow$ & AAE$\downarrow$ \\
    \midrule
    \multicolumn{19}{@{}l}{\emph{Prior-only reference}} \\
    \quad Center Bias
      & 0.679 & 0.906 & 0.095 & 6.635 & 0.110 & 0.185 & 0.162 & 0.215 & 16.07
      & 0.780 & 0.930 & 0.122 & 5.873 & 0.136 & 0.222 & 0.194 & 0.258 & 13.92 \\
    \quad Random Walk
      & 0.623 & 0.824 & 0.096 & 3.926 & 0.101 & 0.147 & 0.090 & 0.413 & 16.34
      & 0.113 & 0.763 & 0.017 & 4.256 & 0.072 & 0.109 & 0.059 & \textbf{0.662} & 14.38 \\
    \midrule
    \multicolumn{19}{@{}l}{\emph{Marginal density estimation}} \\
    \quad GLC~\citep{lai2022glc}
      & 2.216 & 0.953 & 0.279 & 3.288 & 0.249 & 0.421 & 0.348 & 0.531 & 10.21
      & 1.960 & 0.947 & 0.256 & 3.438 & 0.231 & 0.355 & 0.267 & 0.530 & 11.52 \\
    \midrule
    \multicolumn{19}{@{}l}{\emph{Autoregressive}} \\
    
    \quad EgoM2P (zero-shot)~\citep{li2025egom2p}
      & 1.586 & 0.921 & 0.231 & 2.946 & 0.189 & 0.280 & 0.195 & 0.498 & 13.18
      & 1.499 & 0.911 & 0.218 & 3.150 & 0.178 & 0.253 & 0.180 & 0.422 & 13.86 \\
    \quad EgoM2P (LoRA)~\citep{li2025egom2p}
      & 1.735 & 0.921 & 0.247 & 2.737 & 0.198 & 0.293 & 0.205 & 0.515 & 12.81
      & 1.886 & 0.930 & 0.267 & 2.715 & 0.206 & 0.297 & 0.217 & 0.470 & 12.69 \\
      \quad AT~\citep{huang2018predicting}
      & 2.722 & 0.956 & 0.378 & 1.949 & 0.269 & 0.419 & 0.339 & 0.547 & 9.88
      & 2.289 & 0.954 & 0.327 & \textbf{2.156} & 0.197 & 0.346 & 0.268 & 0.487 & 12.01 \\
    \midrule
    \multicolumn{19}{@{}l}{\emph{Flow matching (ours)}} \\
    \quad \textbf{\proj{} (ours)}
      & \textbf{3.534} & \textbf{0.964} & \textbf{0.438} & \textbf{1.936} & \textbf{0.357}
      & \textbf{0.491} & \textbf{0.414} & \textbf{0.603} & \textbf{9.01}
      & \textbf{2.676} & \textbf{0.958} & \textbf{0.358} & 2.213 & \textbf{0.282} & \textbf{0.392} & \textbf{0.307} & 0.540 & \textbf{10.46} \\
    \bottomrule
  \end{tabular}%
  }
\end{table}

\subsection{Efficiency and Accuracy--Cost Trade-off}
\label{app:efficiency}

We measure the cost of \proj{} on a single NVIDIA~H100 (80\,GB) in BF16.
Unless noted, the test clip is a $T{=}64$-frame EGTEA Gaze+ segment, which lasts about $2.7$\,s at $24$\,fps, and we draw $K{=}50$ trajectories with $S{=}50$ Euler steps.

\paragraph{Model size.}
The velocity network has $\sim\!10.4$M parameters, of which $\sim\!7.9$M are in the six DenoiserBlocks of \Sect{sec:method_cfm}.
The spatial-token encoder adds $\sim\!1.6$M, the visual-token projection $\mathbf{W}_{\text{proj}}$ and the learnable-query cross-attention that builds $\mathbf{c}_{\text{task}}$ add $\sim\!0.5$M together, and the coordinate, timestep, and self-conditioning encoders and the output layer add the last $\sim\!0.4$M.
The V-JEPA~2 ViT-L encoder has $\sim\!327.8$M parameters, and we train only its LoRA adapters, which add $\sim\!1.9$M.
In total, we train $\sim\!12.2$M parameters.

\paragraph{Inference cost.}
One forward pass of the velocity network on one $64$-frame trajectory costs $\sim\!100$\,GFLOPs.
With $K{=}50$ samples and $S{=}50$ steps, we call the network $2{,}500$ times, for $\sim\!250$\,TFLOPs per clip.
The encoder runs only once per clip and adds less than $5\%$ to this cost.
In wall-clock time, one clip takes $\sim\!3.37$\,s, of which the encoder with LoRA takes $\sim\!450$\,ms and Euler sampling takes $\sim\!2.92$\,s (mean over five runs), with a peak GPU memory of $9.2$\,GB.

\paragraph{Training time.}
End-to-end training with LoRA takes $\sim\!3$ days on EGTEA Gaze+ and $\sim\!5$ days on Ego4D, both on $4{\times}$\,H100 GPUs in BF16 mixed precision.

\paragraph{Accuracy--cost trade-off.}
\label{app:ks_grid}
Inference cost depends on two settings: $K$, the number of trajectory samples per clip, and $S$, the number of Euler steps per sample.
The main comparison uses a conservative $K{=}50$ and $S{=}50$.
\Tbl{tab:ks_grid} sweeps both with the LoRA model and reports F1 and the inference time per $64$-frame clip on a single H100.
At $K{=}30$ and $S{=}5$, F1 is $0.493$ and inference takes $0.73$\,s, $4.6\times$ faster than the default setting.
We recommend $K$ between $10$ and $30$ with $S{=}5$.

\begin{table}[h]
  \caption{\textbf{F1 and inference time over the sample count $K$ and the Euler step count $S$} (EGTEA Gaze+, \proj{} LoRA). Time is per $64$-frame clip on a single H100. The recommended setting is in bold.}
  \label{tab:ks_grid}
  \centering
  \footnotesize
  \setlength{\tabcolsep}{4pt}
  \begin{tabular}{@{}c ccccc c ccccc@{}}
    \toprule
    & \multicolumn{5}{c}{F1$\uparrow$} & & \multicolumn{5}{c}{Time (s)$\downarrow$} \\
    \cmidrule(lr){2-6} \cmidrule(lr){8-12}
    $K$ \textbackslash{} $S$ & 1 & 2 & 5 & 10 & 50 & & 1 & 2 & 5 & 10 & 50 \\
    \midrule
    1  & 0.453 & 0.465 & 0.444 & 0.427 & 0.405 & & 0.49 & 0.49 & 0.51 & 0.54 & 0.77 \\
    10 & 0.460 & 0.481 & 0.484 & 0.479 & 0.468 & & 0.52 & 0.53 & 0.57 & 0.63 & 1.14 \\
    30 & 0.460 & 0.484 & \textbf{0.493} & 0.492 & 0.486 & & 0.58 & 0.62 & \textbf{0.73} & 0.91 & 2.35 \\
    50 & 0.460 & 0.485 & 0.495 & 0.495 & 0.491 & & 0.65 & 0.71 & 0.89 & 1.21 & 3.37 \\
    \bottomrule
  \end{tabular}
\end{table}

\Tbl{tab:method_time} compares inference time across methods on the same H100.
With the recommended setting, \proj{} is faster than every baseline.

\begin{table}[h]
  \caption{\textbf{Inference time across methods} on a single H100. The ms / frame column gives the time to predict gaze for one video frame, and throughput is its inverse.}
  \label{tab:method_time}
  \centering
  \footnotesize
  \setlength{\tabcolsep}{6pt}
  \begin{tabular}{@{}l cc@{}}
    \toprule
    Method & ms / frame$\downarrow$ & Throughput (fps)$\uparrow$ \\
    \midrule
    GLC~\citep{lai2022glc}           & 12.5 & 80 \\
    AT~\citep{huang2018predicting}   & 219  & 4.6 \\
    EgoM2P~\citep{li2025egom2p}      & 506  & 2.0 \\
    \proj{} ($K{=}50$, $S{=}50$)     & 53   & 19 \\
    \textbf{\proj{} ($K{=}30$, $S{=}5$)} & \textbf{11.4} & \textbf{88} \\
    \bottomrule
  \end{tabular}
\end{table}

\paragraph{Sample count $K$.}
\label{app:k_sample_sweep}
\Tbl{tab:k_sample_sweep} sweeps the sample count $K\in\{1, 5, 10, 30, 50\}$ for \proj{} (LoRA), using the same $K{=}50$ samples as \Tbl{tab:dynamics} and subsampling the first $K$.
The mean over $K$ barely changes ($1.05 \to 1.04$ for ADE, $0.96 \to 0.95$ for DTW).
The gain in best-of-$K$ therefore comes from drawing more samples, not from better individual samples.
Best-of-$K$ changes little after $K{=}30$ (ADE $0.64$ vs.\ $0.60$ at $K{=}50$), so $K{=}50$ in the main comparison is a safe choice.

\Tbl{tab:k_sample_sweep} also reports the per-frame metrics for the same $K$ values.
They also rise with $K$ and flatten out: $K{=}10$ keeps $96\%$ of the F1 at $K{=}50$, and $K{=}30$ is close to $K{=}50$ on every metric.

\begin{table}[h]
  \caption{\textbf{Sample count $K$ sweep on EGTEA Gaze+.} ADE and DTW are reported in $\times 10^{2}$\,px following \Tbl{tab:dynamics}; ``Mean'' and ``Best'' are mean-over-$K$ and best-of-$K$. At $K{=}1$ the two coincide. The $K{=}50$ row matches \Tbl{tab:main} and \Tbl{tab:dynamics}.}
  \label{tab:k_sample_sweep}
  \centering
  \footnotesize
  \setlength{\tabcolsep}{4pt}
  \renewcommand{\arraystretch}{0.95}
  \begin{tabular}{@{}c ccccc cc cc@{}}
    \toprule
    & & & & & & \multicolumn{2}{c}{ADE ($\times 10^{2}$\,px)$\downarrow$} & \multicolumn{2}{c}{DTW ($\times 10^{2}$\,px)$\downarrow$} \\
    \cmidrule(lr){7-8} \cmidrule(lr){9-10}
    $K$ & AUC$\uparrow$ & F1$\uparrow$ & P$\uparrow$ & R$\uparrow$ & AAE$\downarrow$ & Mean & Best & Mean & Best \\
    \midrule
    1   & 0.866 & 0.398 & 0.350 & 0.462 & 11.10 & 1.05 & 1.05 & 0.96 & 0.96 \\
    5   & 0.942 & 0.457 & 0.378 & 0.578 & 9.61  & 1.05 & 0.80 & 0.95 & 0.68 \\
    10  & 0.953 & 0.469 & 0.395 & 0.579 & 9.28  & 1.05 & 0.73 & 0.95 & 0.60 \\
    30  & 0.961 & 0.487 & 0.399 & 0.623 & 9.05  & 1.04 & 0.64 & 0.95 & 0.51 \\
    50  & 0.964 & 0.491 & 0.414 & 0.603 & 9.01  & 1.04 & 0.60 & 0.95 & 0.48 \\
    \bottomrule
  \end{tabular}
\end{table}

\subsection{Capacity Sweep over Task Queries}
\label{app:k_sweep}

Table~\ref{tab:k_sweep} reports the full sweep over $M \in \{1,2,4,8,16\}$, the number of learnable task queries, with the global pool token kept (\Sect{sec:design}).
$M{=}4$ gives the best AUC and F1.

\begin{table}[h]
  \caption{\textbf{Capacity sweep over learnable task queries.} $M$ is the number of learnable task queries, and every variant keeps the global pool token. Per-column best in bold.}
  \label{tab:k_sweep}
  \centering
  \footnotesize
  \setlength{\tabcolsep}{4pt}
  \renewcommand{\arraystretch}{0.95}
  \begin{tabular}{@{}c ccccc@{}}
    \toprule
    $M$ & AUC$\uparrow$ & F1$\uparrow$ & P$\uparrow$ & R$\uparrow$ & AAE$\downarrow$ \\
    \midrule
    1            & 0.968          & 0.492          & \textbf{0.404} & 0.630          & \textbf{8.68} \\
    2            & 0.967          & 0.486          & 0.392          & 0.639          & 8.87 \\
    \textbf{4}   & \textbf{0.970} & \textbf{0.493} & 0.401          & 0.638          & 8.81 \\
    8            & 0.968          & 0.483          & 0.390          & 0.635          & 8.79 \\
    16           & 0.968          & 0.480          & 0.381          & \textbf{0.650} & 8.75 \\
    \bottomrule
  \end{tabular}
\end{table}

\subsection{Additional Ablations}
\label{app:ablations_extra}

\paragraph{Causal variant.}
\label{app:causal}
We train a causal variant of \proj{} that sees only past and current frames.
It adds a causal mask to the time-aware self-attention, builds the task-token bank from past and current frames, and extracts visual features with a causal sliding window.
It loses at most $12.0\%$ across the five metrics (\Tbl{tab:causal}).

\paragraph{Flow matching vs.\ diffusion.}
\label{app:fm_vs_diffusion}
We train a diffusion version of \proj{} that changes only the objective (noise prediction, cosine schedule, DDIM sampling).
Both heads call the same network once per step, so they take the same time at each step count $S$.
As \Tbl{tab:fm_vs_diffusion} shows, flow matching is best at $S{=}5$, while diffusion needs all $50$ steps, $3.8\times$ the time, and is still less accurate.

\begin{table}[h]
  \begin{minipage}[t]{0.47\linewidth}
    \centering
    \caption{\textbf{Bidirectional vs.\ causal \proj{}.} The causal variant sees only past and current frames.}
    \label{tab:causal}
    \scriptsize
    \setlength{\tabcolsep}{3pt}
    \begin{tabular}{@{}l ccccc@{}}
      \toprule
      Variant & AUC$\uparrow$ & F1$\uparrow$ & P$\uparrow$ & R$\uparrow$ & AAE$\downarrow$ \\
      \midrule
      Bidirectional & 0.970 & 0.493 & 0.401 & 0.638 & 8.81 \\
      Causal        & 0.953 & 0.437 & 0.353 & 0.574 & 8.71 \\
      \bottomrule
    \end{tabular}
  \end{minipage}\hfill
  \begin{minipage}[t]{0.5\linewidth}
    \centering
    \caption{\textbf{Flow matching vs.\ diffusion over the number of sampling steps $S$} ($K{=}50$). Time is per $64$-frame clip on a single H100.}
    \label{tab:fm_vs_diffusion}
    \scriptsize
    \setlength{\tabcolsep}{3pt}
    \begin{tabular}{@{}l cccc@{}}
      \toprule
      $S$ & 2 & 5 & 10 & 50 \\
      \midrule
      F1$\uparrow$, flow matching (ours) & 0.496 & \textbf{0.506} & 0.503 & 0.493 \\
      F1$\uparrow$, diffusion            & 0.072 & 0.359 & 0.462 & 0.485 \\
      Time (s)$\downarrow$               & 0.71  & 0.89  & 1.21  & 3.37 \\
      \bottomrule
    \end{tabular}
  \end{minipage}
\end{table}

\end{document}